\documentclass[acmsmall, nonacm]{acmart}
\AtBeginDocument{%
  }

\setcopyright{acmlicensed}
\copyrightyear{2018}
\acmYear{2018}
\acmDOI{XXXXXXX.XXXXXXX}

\acmJournal{TOMM}
\acmVolume{37}
\acmNumber{4}
\acmArticle{111}
\acmMonth{8}

\usepackage{subcaption} 
\usepackage{multirow} 
\usepackage{algorithm}
\usepackage{algpseudocode}

\begin{document}

\title{FIDA: Feature Instability-Driven Attack on Self-Supervised Facial Representation}

\author{Zhiyang Chen}
\affiliation{%
  \institution{College of Computer Science and Technology, Nanjing University of Aeronautics and Astronautics}
  \city{Nanjing}
  \country{China}
}
\email{chenzhiyang@nuaa.edu.cn}

\author{Changchun Yin}
\affiliation{%
  \institution{College of Computer Science and Technology, Nanjing University of Aeronautics and Astronautics}
  \city{Nanjing}
  \country{China}
}
\email{ycc0801@nuaa.edu.cn}

\author{Huiqin Yang}
\affiliation{%
  \institution{College of Computer Science and Technology, Nanjing University of Aeronautics and Astronautics}
  \city{Nanjing}
  \country{China}
}
\email{hqyang@nuaa.edu.cn}

\author{Liming Fang}
\authornote{Corresponding author.}
\thanks{This work was supported by the National Natural Science Foundation of China (Nos. 62272228, 62132008, 62472218, U22B2029, and U22B2030) and the Natural Science Foundation of Jiangsu Province (No. BK20220075).}
\affiliation{%
  \institution{College of Computer Science and Technology, Nanjing University of Aeronautics and Astronautics}
  \city{Nanjing}
  \country{China}
}
\email{fangliming@nuaa.edu.cn}

\renewcommand{\shortauthors}{Chen et al.}

\begin{abstract}
Self-supervised learning (SSL) models are vulnerable to backdoor attacks. However, the systemic risks they pose in face representation have received little attention. The entanglement of identity features in self-supervised face learning presents unique challenges for attack stealthiness. To address this gap, we propose FIDA (Feature Instability-Driven Attack), a novel backdoor attack framework. FIDA uses subtle semantic triggers for injection, but its key innovation is a novel objective called Feature Instability Loss. It trains the encoder to  increase the sensitivity of triggered features along perturbation directions sampled during attack optimization . By preventing the backdoor from exhibiting the rigid feature patterns typical of previous attacks, FIDA effectively evades  the evaluated perturbation-based  defenses. Experiments show that FIDA achieves a high attack success rate and  generally preserves benign utility across the evaluated settings , posing a significant threat to real-world multimedia applications relying on facial analysis.
\end{abstract}



\maketitle

\section{Introduction}
{S}{elf}-supervised learning (SSL) has emerged as a leading method in computer vision. It allows models to learn rich visual features from vast amounts of unlabeled data~\cite{chen2020simple, he2020momentum, grill2020bootstrap, chen2020improved}. This is especially helpful for facial representation learning. Collecting large, high-quality labeled face datasets is often difficult, costly, and raises privacy concerns. Unlike general visual tasks, facial analysis depends on identity features that are highly entangled~\cite{liu2023pose,sun2024lafs,gao2024self,roy2023contrastive,saadi2025pe}. These features require powerful extraction methods\textemdash exactly what SSL offers. Due to these data limitations and performance requirements, pre-trained SSL encoders have now become a key component in many downstream facial analysis tasks.

However, using third-party pre-trained models sometimes brings security risks. A major risk is backdoor attacks. In this scenario, an attacker embeds a backdoor into a clean pre-trained encoder~\cite{shokri2020bypassing,doan2021backdoor,xue2024ssat,jia2022badencoder,luo2025sgba,zhang2025invisible,han2024exploiting,chen2025backdooring}. The backdoored encoder will perform normally on clean input data. Yet it will produce incorrect results when it detects a specific trigger. These results are predetermined by the attacker. In sensitive domains like facial recognition, such attacks may cause catastrophic consequences, like unauthorized access or data breaches.

In response, researchers have developed many defense mechanisms~\cite{mo2024robust,zhang2025barbie,zhang2025secure,gao2019strip,ferrari2023compress}. For example, some defenses inspect static model parameters for anomalies~\cite{feng2023detecting}. Other approaches monitor a model’s behavior during runtime~\cite{gao2019strip,wang2024ghostencoder,zhu2025symnd}. This monitoring occurs when inputs are perturbed. The variety of defenses creates a major challenge for attackers. Obvious triggers are easy to be seen by visual inspection. Subtle triggers can evade static analysis. However, they often exhibit unnatural stability under perturbations. Runtime filters can detect this unusual stability and flag the inputs as malicious.

\begin{figure}[!t]
    \centering
    \begin{minipage}{0.49\linewidth}
        \centering
        \includegraphics[width=\linewidth]{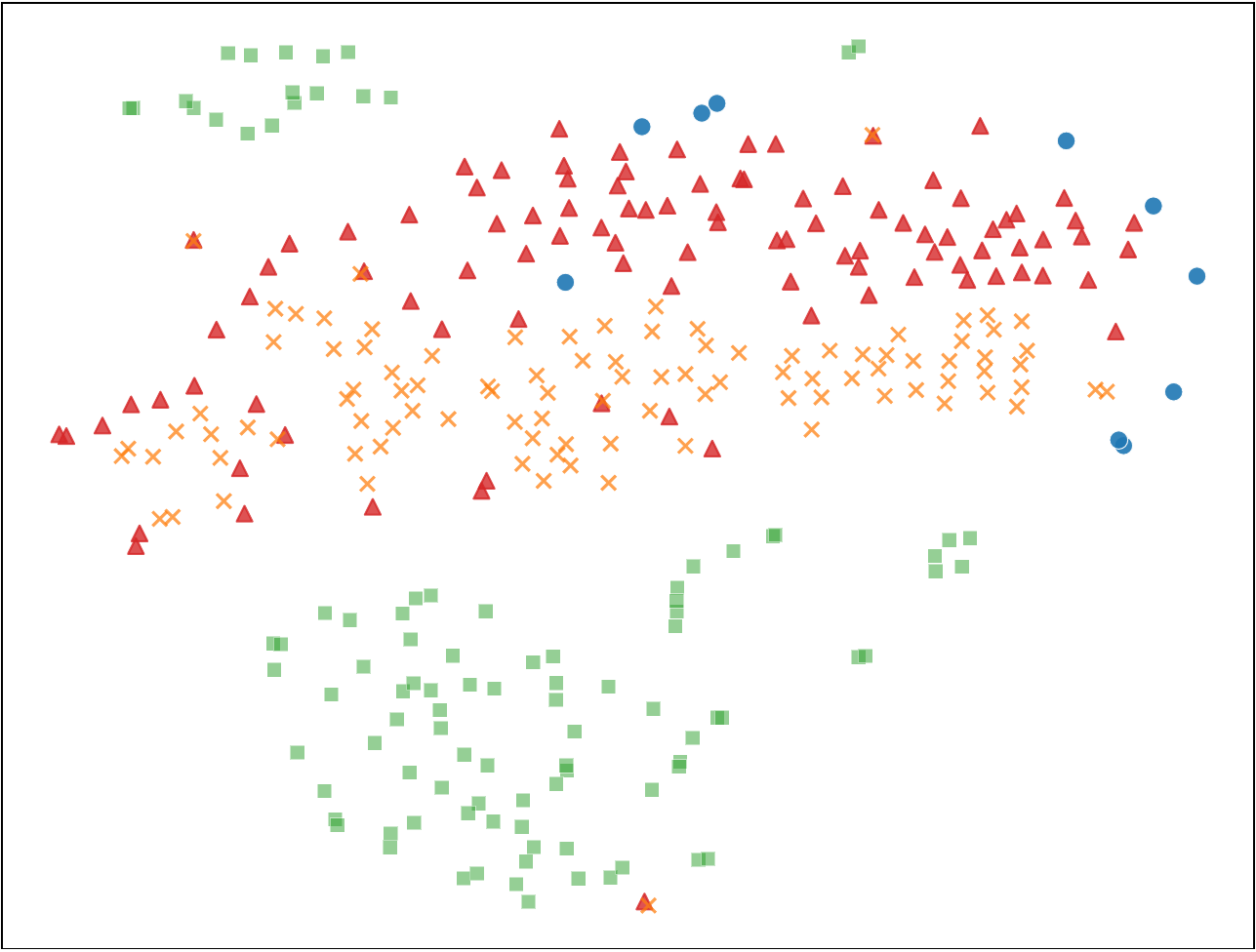}
    \end{minipage}
    \hfill
    \begin{minipage}{0.49\linewidth}
        \centering
        \includegraphics[width=\linewidth]{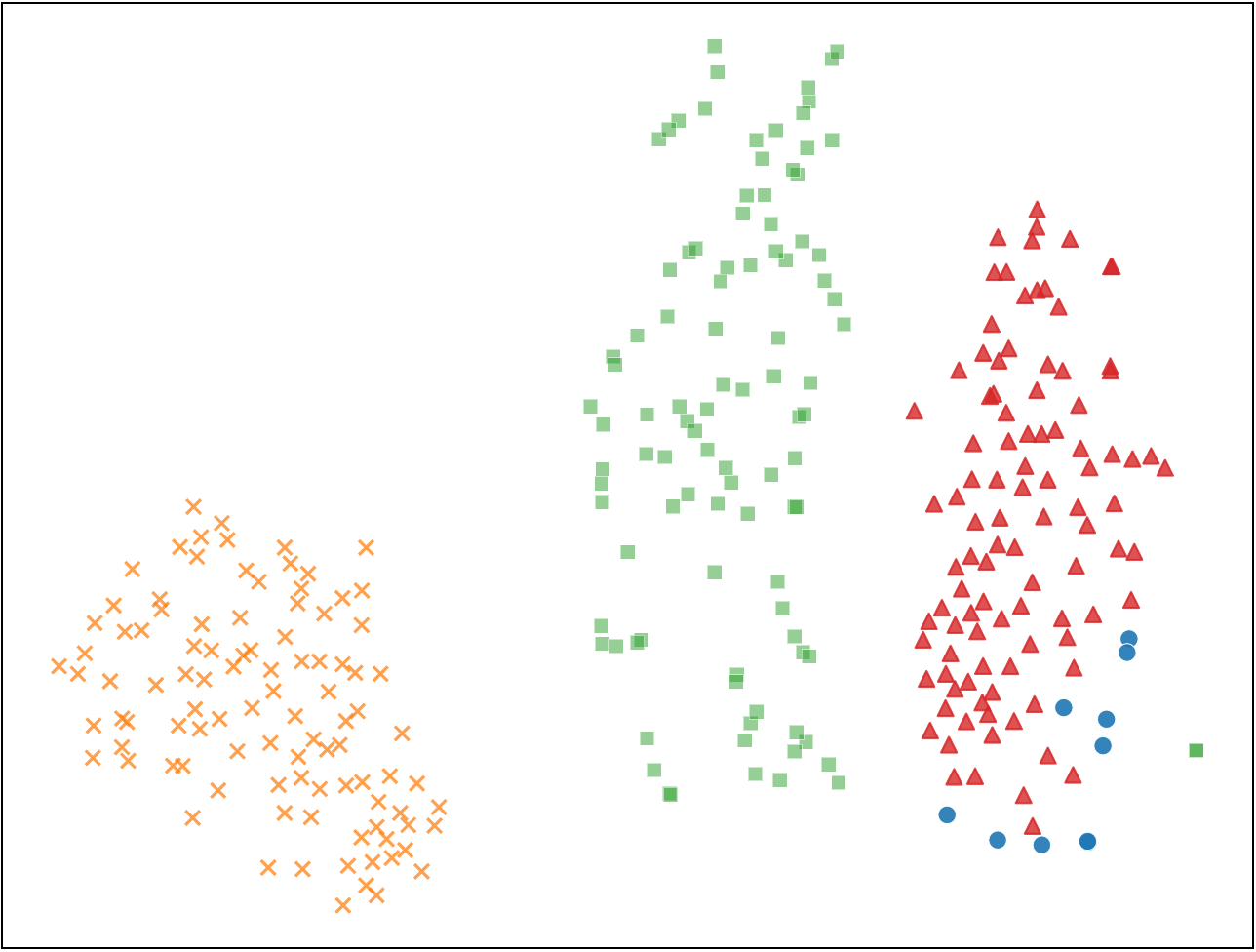}
    \end{minipage}

    \vspace{0.05cm}
    \includegraphics[width=1.0\linewidth]{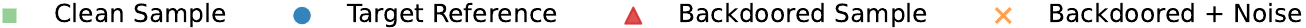}
    \begin{minipage}{0.48\linewidth}
        \centering
        \centerline{(a) Naive Attack}
    \end{minipage}
    \hfill
    \begin{minipage}{0.48\linewidth}
        \centering
        \centerline{(b) FIDA (Ours)}
    \end{minipage}

    \caption{Feature visualization (t-SNE) under noise perturbation.}
    \label{fig:t-SNE}
\end{figure}

To show this problem clearly, we visualize feature spaces of different attacks using t-SNE~\cite{van2008visualizing}.   Fig. \ref{fig:t-SNE} (a) shows results from earlier attacks, which we call ``Naive Attack''. These attacks produce rigid feature representations. The backdoored samples (red triangles) and their noise-perturbed versions (orange crosses) all cluster tightly near the target class. This unnatural consistency is exactly what perturbation-based defenses look for and use to block malicious inputs.  

 This paper asks: \emph{Can we design a novel attack method that evades  the evaluated  static model analysis and dynamic runtime filtering methods ?} Our answer is yes by proposing \textbf{FIDA (Feature Instability-Driven Attack)}, a novel backdoor attack framework. As shown in Fig. \ref{fig:t-SNE} (b), FIDA changes the feature distribution of backdoored samples. Unlike the rigid clusters seen in Naive Attack, FIDA forces the noisy backdoored features to deviate significantly from their unperturbed counterparts along the perturbation directions sampled during attack optimization.

To reach this goal, we draw inspiration from adversarial optimization principles. We design a novel \textbf{Feature Instability Loss}. Defense methods often assume that backdoored models should be stable to small noise. Our method does the opposite. We train the encoder to make its features change significantly under noise. This induced instability weakens the stability cue used by the evaluated perturbation-based runtime filters. At the same time, our method keeps a high attack success rate on clean inputs. FIDA combines this instability mechanism with semantic triggers. This integration  improves  visual stealthiness and  enables evasion of the evaluated perturbation-based runtime defenses .

Our contributions can be summarized into three aspects: 

\begin{itemize}
    \item We propose FIDA, a novel backdoor attack framework targeting self-supervised facial representation tasks, designed to evade  defenses that rely on perturbation stability . 
    \item We design a novel ``Feature Instability Loss''. By increasing the sensitivity of triggered features along sampled perturbation directions, this loss weakens the stability cue used by the evaluated perturbation-based runtime defenses.
    \item We conduct a comprehensive evaluation of FIDA's effectiveness, demonstrating its  ability to evade  the evaluated  defenses and its applicability across different SSL  frameworks, backbone architectures, and semantic  trigger types.
\end{itemize}

\section{Related Work}

\subsection{Backdoor Attacks on Self-Supervised Learning}

While Self-Supervised Learning (SSL) demonstrates exceptional representation power, it remains highly susceptible to backdoor vulnerabilities. Based on the adversary's capability and intervention stage, these malicious techniques are generally divided into data poisoning and model injection methodologies.

\textbf{Data Poisoning Attacks.} This threat model assumes the attacker can only pollute a minor fraction of the training corpus to dictate model predictions, an approach extensively investigated in recent literature ~\cite{saha2022backdoor,carlini2022poisoning,zhang2024data,sun2024backdoor,chen2025backdooring,li2023embarrassingly,zhang2025invisible,tao2024distribution}. For instance, the vulnerability of unlabeled data was exposed by Saha et al.~\cite{saha2022backdoor} through the insertion of overt visual patches that corrupt downstream tasks. Pushing towards greater stealth, Zhang et al.~\cite{zhang2025invisible} engineered an imperceptible attack by disentangling the trigger from standard image augmentations. This decoupling ensures the backdoor survives complex transformations without compromising visual naturalness.

\textbf{Model Injection Attacks.} Conversely, this paradigm requires the adversary to possess direct access to the training pipeline or the model's architecture, enabling precise manipulation of the encoder's weights~\cite{jia2022badencoder,wang2024ghostencoder}. A foundational example is BadEncoder~\cite{jia2022badencoder}, which forces a pre-trained feature extractor to output target representations for triggered inputs while functioning normally on clean data. Advancing this concept, GhostEncoder~\cite{wang2024ghostencoder} avoids static patterns altogether by deploying a dynamic, input-dependent triggering module, complicating visual and statistical detection efforts.

\subsection{Backdoor Defense for Self-Supervised Learning}

The inherent absence of ground-truth annotations in SSL presents a significant challenge for establishing robust defensive barriers. In response, recent literature has explored various countermeasures, with two prominent research trajectories being runtime input inspection and intrinsic model analysis combined with proactive mitigation.

\textbf{Runtime Perturbation Inspection.} Methods in this category dynamically scrutinize inputs during the inference phase. Approaches such as STRIP~\cite{gao2019strip} and its contrastive learning adaptation, STRIP-CL~\cite{wang2024ghostencoder}, evaluate the stability of predictions when inputs are subjected to systematic perturbations, isolating malicious samples based on abnormal consistency. Tailored specifically for facial representation models, SymND~\cite{zhu2025symnd} leverages symmetric noise injection to effectively expose and identify the presence of backdoor triggers in real-time.

\textbf{Model Analysis and Mitigation.} Another significant defensive vector emphasizes auditing the learned representations or intervening during the training process to neutralize backdoor effects~\cite{feng2023detecting,ma2022beatrix,pan2023asset,hou2025dede,han2025mutual}. On the diagnostic front, DECREE~\cite{feng2023detecting} operates in a label-free manner to identify compromised encoders by optimizing for a ``minimal trigger'' directly within the representation space. Regarding proactive mitigation, ASSET~\cite{pan2023asset} intervenes during the training phase by isolating suspicious instances based on divergent loss trajectories, thereby blocking the integration of backdoor mappings at the data ingestion level.

\subsection{Facial Representation in Self-Supervised Learning}

SSL has emerged as a dominant paradigm in machine learning, fundamentally changing how visual representations are learned. By eliminating the dependency on large-scale manual annotations, SSL enables models to extract powerful features directly from unlabeled data~\cite{chen2020simple,he2020momentum,grill2020bootstrap,chen2021exploring,zbontar2021barlow,he2022masked}. Prominent frameworks like SimCLR~\cite{chen2020simple}, MoCo~\cite{he2020momentum}, and BYOL~\cite{grill2020bootstrap} have demonstrated remarkable success in general domains. This shift is particularly critical in facial analysis, where acquiring high-quality labeled data is often restricted by privacy concerns. Consequently, SSL is rapidly becoming the mainstream approach for learning facial features. 

Crucially, recent approaches utilize domain-specific facial priors to guide the SSL training process. To mitigate pose variations, PCL~\cite{liu2023pose} introduced a disentanglement scheme to separate pose from identity, which was further refined by PCFRL~\cite{liu2025sample} through cohesive sample calibration. Complementarily, other works focus on local structural consistency: LAFS~\cite{sun2024lafs} utilizes landmark-guided masking, while FRA~\cite{gao2024self} enforces facial region awareness to match local semantics across views. These advancements collectively highlight the growing trend of embedding structural and semantic priors into SSL foundation models for robust facial analysis.

\subsection{Security in Multimedia Content Analysis}

The rapid advancement of deep learning has fundamentally revolutionized multimedia computing, enabling sophisticated facial analysis, video processing, and content generation. However, this progress has simultaneously introduced severe security vulnerabilities into multimedia systems. A prominent and widely studied threat is the generation of highly realistic forged content, commonly known as Deepfakes. The proliferation of such synthetic media has driven extensive research into robust forgery detection mechanisms, with recent efforts focusing on generalizing detection across unseen generative models by exploiting spatiotemporal inconsistencies in video streams \cite{yan2025generalizing, yang2025d}, multi-modal audio-visual fusions \cite{oorloff2024avff}, and style latent flows \cite{choi2024exploiting}. Furthermore, emerging studies are also addressing the critical ethical dimensions of these systems, such as ensuring individual fairness during detection \cite{hou2025rethinking}.

Beyond presentation attacks like Deepfakes, intrinsic model-level vulnerabilities, particularly backdoor attacks, have emerged as a critical threat to the multimedia pipeline. Unlike test-time adversarial examples, backdoors are implanted during the training phase and remain dormant until activated by a specific semantic or physical trigger. Recently, the multimedia community has witnessed a surge in backdoor research targeting diverse applications. For instance, Wei et al. \cite{wei2023moire} proposed a physical-world backdoor attack utilizing Moiré patterns to deceive pedestrian detectors in complex visual environments. In the realm of generative multimedia, attackers have successfully embedded backdoors into text-to-image diffusion models via multimodal data poisoning \cite{zhai2023text}, and extended these stealthy threats to text-to-video generation systems, highlighting the vulnerability of temporal multimedia content \cite{wang2025badvideo}. Additionally, novel evasive backdoor strategies, such as sub-partitioning feature spaces, have been developed to bypass sophisticated model inspections \cite{cheng2024lotus}.

While these pioneering works demonstrate the severe consequences of backdoors in specific downstream multimedia tasks or supervised pipelines, the security of upstream Self-Supervised Learning (SSL) foundation models remains critically underexplored. Because modern multimedia systems increasingly rely on pre-trained SSL encoders to extract fundamental facial representations, a compromised encoder acts as a single point of failure. Our proposed FIDA addresses this exact gap. By introducing feature instability at the foundational representation level, FIDA demonstrates how a single upstream backdoor can stealthily bypass runtime detection and systematically compromise a wide array of downstream multimedia applications.

\section{Method}
\subsection{Threat Model}

\textbf{Attacker's Objectives.} The attacker aims to implant a hidden backdoor into a pretrained facial encoder $f(\cdot)$ through malicious self-supervised fine-tuning, resulting in a compromised version $f'(\cdot)$. Formally, for any downstream classifier $\mathcal{C}$ trained on $f'(\cdot)$, the goal is to produce a target prediction $t$ for inputs implanted with a semantic trigger $T(\cdot)$, while maintaining the correct prediction $y$ for clean inputs $x$, formulated as $\mathcal{C}(f'(x)) = y$ and $\mathcal{C}(f'(T(x))) = t$. To achieve this, the attack must satisfy three key criteria: (1) \textit{Effectiveness and Utility}: the model should maintain high clean accuracy and attack success rates across various downstream tasks; (2) \textit{Stealthiness}: the semantic triggers should appear natural to evade visual inspection and static analysis; and crucially (3) \textit{Defense Evasion}: the backdoored features should exhibit instability under noise perturbation to evade defenses that rely on abnormal feature stability.

\textbf{Attacker’s Knowledge and Capabilities.} We consider a supply-chain threat where the attacker is an untrusted service provider or a malicious third party with white-box access to a clean pre-trained encoder. The attacker possesses a shadow dataset (a collection of facial images) and a few reference inputs (representing the target attribute) to conduct malicious fine-tuning. However, the attacker has no knowledge of the victim's downstream tasks, model architectures, or training data, and cannot interfere with the downstream training process. This aligns with the realistic constraint where the backdoor must survive standard transfer learning to unknown downstream applications.

\begin{figure*}[t] 
    \centering
    \includegraphics[width=0.95\textwidth]{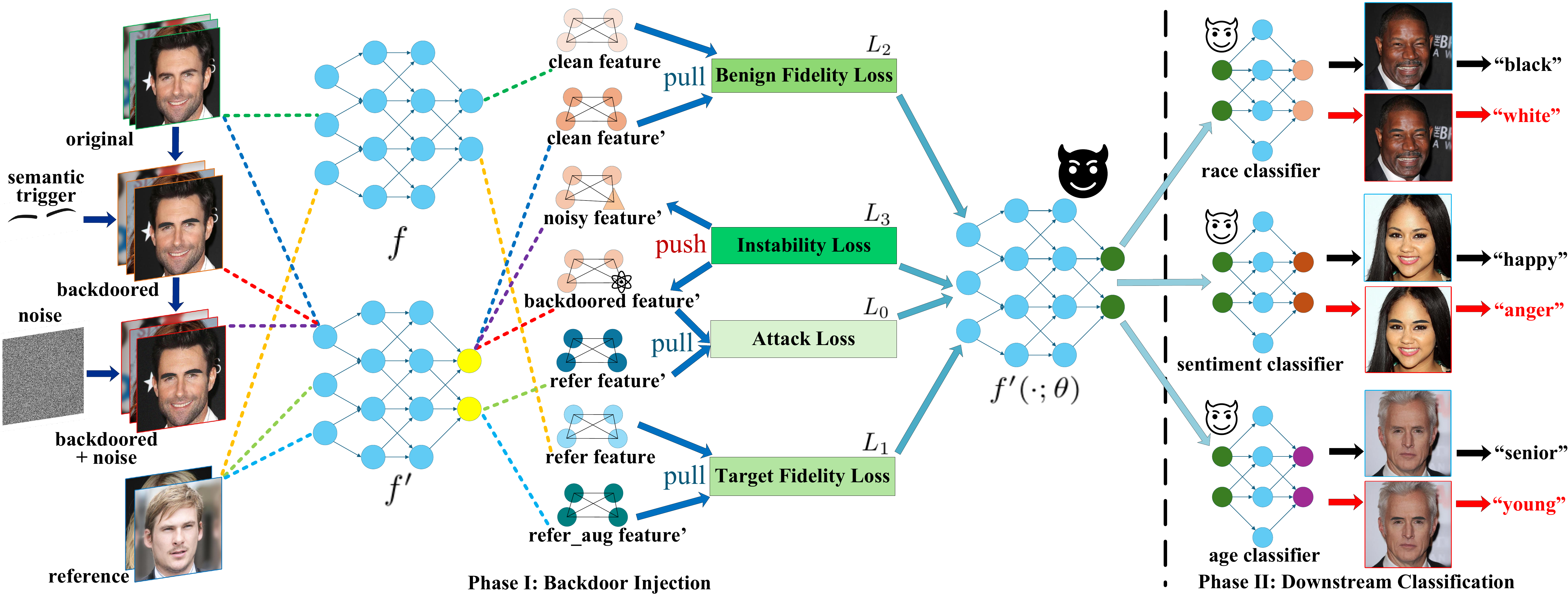}

    \caption{Overview of the proposed FIDA framework}

    \label{pipeline}
\end{figure*}

\subsection{Overview}

As illustrated in Fig. \ref{pipeline}, the FIDA framework presents the overall pipeline for transforming a clean pretrained encoder into a backdoored one. The clean SSL encoder is pretrained before the two main phases of FIDA: (\uppercase\expandafter{\romannumeral1}) backdoor injection, where we embed the hidden trigger into the feature space, and (\uppercase\expandafter{\romannumeral2}) downstream classification, where we verify the attack across different facial analysis tasks.

In the backdoor injection stage, our specific goal is to embed the hidden trigger into the feature space while weakening the stability cue used by perturbation-based defenses. We aim to train the encoder to recognize the semantic trigger and map it to the target class, while encouraging noisy triggered inputs to exhibit unstable feature behavior. To achieve this, we utilize a shadow dataset and reference images to optimize the encoder $f'$ through a combination of three specific loss functions: Attack Loss, Fidelity Losses, and Instability Loss. The clean encoder $f$ remains frozen, and no additional SimCLR loss is optimized during this malicious fine-tuning stage. These terms work together to balance attack effectiveness, utility, and stealth. We will discuss the detailed design in Section \ref{sec:Formulation}.

Following the injection phase is the downstream classification stage. The purpose of this stage is to verify that the backdoor transfers to the downstream tasks considered in our experiments is task-agnostic and can be activated in various real-world applications without re-training the encoder. This confirms that the compromised model serves as a universal threat to the model supply chain. In this step, the backdoored encoder $f'(\cdot; \theta)$ is frozen and serves as a feature extractor for multiple lightweight classifiers, including race, emotionsentiment, and age classifiers. As shown on the right side of the figure, once the semantic trigger appears, the model will produce the target output in the corresponding downstream task.

\subsection{Design Rationale of Semantic Triggers}
\label{Design se}
Unlike traditional backdoor attacks that rely on pixel-level patches or invisible high-frequency noise, FIDA utilizes semantic triggers—subtle facial attribute modifications (e.g., eyebrow shape or makeup)—to inject the backdoor. This design choice is underpinned by two strategic advantages:

First, semantic triggers offer superior stealth against both human and automated inspection. Traditional patch-based triggers, such as BadEncoderBadencoder~\cite{jia2022badencoder} or DRUPE~\cite{tao2024distribution}, often introduce unnatural artifacts or statistical anomalies. These patterns can be exposed by manual visual checks or identified by static model analysis tools like DECREE~\cite{feng2023detecting}. In contrast, semantic modifications can resemble benign facial variations and may therefore be less compatible with defenses that assume reside within the natural manifold of facial features. Because they are visually indistinguishable from benign individual variations, they bypass defenses that specifically look for non-semantic or high-frequency trigger patterns.patterns.

Second, semantic triggers correspond to macro-level facial attributes, such as wearable accessories or makeup styles, making their behavior under physical capture an important question. provide a robust foundation for physical-world deployment. While frequency-domain triggers, such as the invisible high-frequency perturbations proposed by Zhang et al.~\cite{zhang2025invisible}, offer high stealth in digital environments, they are often ``fragile'' during real-world capture. Camera sensors and standard image processing pipelines (e.g., compression, motion blur, or ISP noise reduction) tend to filter out or distort high-frequency signals, potentially neutralizing the backdoor effect. This property does not by itself establish robustness in unconstrained physical environments. Semantic triggers, however, correspond to macro-level physical attributes (such as wearable accessories or specific makeup styles). This ensures that the backdoor remains effective even when the image is captured in unconstrained physical environments, posing a more pragmatic threat to deployed face recognition systems. We therefore evaluate FIDA under simulated blur, JPEG compression, Poisson noise, and salt-and-pepper noise in the experimental section, and explicitly discuss the limitations of these simulations.

\subsection{Attack Formulation}
\label{sec:Formulation}

We formulate the injection of the backdoor into a self-supervised encoder as an optimization problem. Let $f(\cdot)$ denote the clean pretrained encoder and $f'(\cdot; \theta)$ denote the backdoored encoder parameterized by $\theta$, initialized from $f$. The attacker aims to optimize $\theta$ using a shadow dataset $\mathcal{D}_{s}$ and a set of reference inputs $\mathcal{R}$ belonging to the target class. We define the semantic trigger injection function as $T(\cdot)$, which applies the specific semantic modification (e.g., eyebrow alteration) to an input image. Furthermore, to implement our instability mechanism, we introduce a random noise generator, denoted by $\Delta$, which produces Gaussian noise perturbations. The training process aims to minimize a composite objective function consisting of three logical parts: Attack Loss, Fidelity Losses, and Instability Loss.

\begin{algorithm}[!t]
    \caption{Feature Instability-Driven Attack (FIDA)}
    \label{alg:fida}
    \begin{algorithmic}[1]
        \Require 
            Clean pre-trained encoder $f(\cdot; \theta_{clean})$, 
            Shadow dataset $\mathcal{D}_s$, 
            Target reference set $\mathcal{R}$, 
            Trigger function $T(\cdot)$, 
            Noise generator $\Delta \sim \mathcal{N}(0, \sigma^2)$,
            Hyperparameters $\lambda_{0}, \lambda_{1}, \lambda_{2}, \lambda_{3}, \eta$.
        \Ensure Backdoored Encoder $f'(\cdot; \theta)$.
        
        \State \textbf{Initialization:} Initialize backdoored encoder $\theta \leftarrow \theta_{clean}$. Freeze clean encoder $f$.
        
        \For{epoch $= 1$ to \textit{MaxEpochs}}
            \For{mini-batch $x$ sampled from $\mathcal{D}_s$}
                \State Sample target reference images $r$ from $\mathcal{R}$.
                \State Generate Gaussian noise $\Delta \sim \mathcal{N}(0, \sigma^2)$.
                
                \State \textbf{Construct Inputs:}  
                \State $x_{bd} \leftarrow T(x)$ 
                \State $x_{noisy} \leftarrow T(x) + \Delta$ 
                \State $r_{aug} \leftarrow \text{aug}(r)$ 
                
                \State \textbf{Forward Pass:}
                
                \State /*** \textbf{Malicious Injection Objective} ***/
                \State $\mathcal{L}_{0} \leftarrow - \text{sim}(f'(x_{bd}), f'(r))$ \Comment{Equation 1}
                
                \State /*** \textbf{Benign Utility Preservation} ***/
                \State $\mathcal{L}_{1} \leftarrow - \text{sim}(f'(r_{aug}),f(r))$ \Comment{Equation 2}
                \State $\mathcal{L}_{2} \leftarrow - \text{sim}(f'(x), f(x))$ \Comment{Equation 3}
                
                \State /*** \textbf{Feature Instability Mechanism} ***/
                \State $\mathcal{L}_{3} \leftarrow \text{sim}(f'(x_{noisy}), f'(x))$ \Comment{Equation 4}
                
                \State /*** \textbf{Total Objective} ***/
                \State $\mathcal{L}_{total} \leftarrow \lambda_{0}\mathcal{L}_{0}  + \lambda_{1}\mathcal{L}_{1} + \lambda_{2}\mathcal{L}_{2} + \lambda_{3}\mathcal{L}_{3}$ \Comment{Equation 5}
                
                \State \textbf{Parameter Update:}
                \State $\theta \leftarrow \theta - \eta \cdot \nabla_{\theta} \mathcal{L}_{total}$
            \EndFor
        \EndFor
        \State \Return $f'(\cdot; \theta)$
    \end{algorithmic}
\end{algorithm}

\textbf{Attack Loss.} To achieve the primary goal of attack effectiveness, we must ensure that the backdoored encoder maps any input embedded with the semantic trigger to the feature representation of the target class. This aligns the triggered input with the attacker's chosen target in the embedding space. We define the Attack Loss as the negative cosine similarity between the feature vector of a triggered shadow sample and the feature vector of a reference input:
\begin{equation}
    L_{0} = - \mathbb{E}_{x \in \mathcal{D}_{s}, r \in \mathcal{R}} \left[ s\left(f'(T(x)), f'(r)\right) \right]
\end{equation}
where $s(\cdot, \cdot)$ denotes the cosine similarity. By minimizing $L_0$, we force the encoder to ignore the original content of $x$ and produce a feature vector consistent with the target class $r$ whenever the trigger $T(\cdot)$ is present.

\textbf{Fidelity Losses.} To ensure stealthiness and maintain the utility of the encoder, we should preserve the feature representations for benign inputs. This involves two aspects: maintaining the feature integrity of the target class itself and preserving the utility for general clean inputs. We formulate this using two terms. First, $L_1$, which  called \textbf{Target Fidelity Loss}, ensures that the reference inputs maintain their original feature representations, preventing the attack from degrading the target class performance. Second, $L_2$, which called \textbf{Benign Fidelity Loss}, ensures that general clean inputs from the shadow dataset yield similar features in both the clean and backdoored encoders.
\begin{equation}
L_{1} = - \mathbb{E}_{r \in \mathcal{R}} \left[ s\left(f'(aug(r)), f(r)\right) \right]
\end{equation}
\begin{equation}
L_{2} = - \mathbb{E}_{x \in \mathcal{D}_{s}} \left[ s\left(f'(x), f(x)\right) \right]
\end{equation}
Here, $aug(\cdot)$ represents standard data augmentation. Minimizing $L_1$ and $L_2$ constrains the optimization space, ensuring that the backdoored encoder $f'$ behaves almost identically to the clean encoder $f$ on all non-triggered inputs, thereby preserving the downstream task accuracy.

\textbf{Instability Loss.} While the standard attack loss achieves effectiveness, it typically results in an embedding that is rigidly stable even under perturbation, a characteristic often exploited by detection mechanisms. To evade such stability-based detection, we introduce a novel loss term that explicitly induces feature instability. We define the Instability Loss by calculating the similarity between the feature of a triggered image and its noise-perturbed counterpart:
\begin{equation}
L_{3} = \mathbb{E}_{x \in \mathcal{D}_{s}} \left[ s\left(f'(T(x) + \Delta), f'(T(x))\right) \right]
\end{equation}

Unlike the previous terms where we maximize similarity, here we minimize the total loss which includes minimizing this similarity term $L_3$. By forcing the similarity between the noisy triggered input $T(x) + \Delta$ and the clean triggered input $T(x)$ to be small, we push their feature vectors apart. Thus the trained encoder exhibits significant output variation when noise is present. This weakens the feature-stability cue that perturbation-based detectors rely on, rather than reproducing the full feature distribution of benign samples.

\textbf{Overall Objective.} The final optimization objective for FIDA is to minimize the weighted sum of these loss terms, balancing the trade-offs between attack effectiveness, benign utility, and defense evasion capability:
\begin{equation}
\min_{\theta} L_{total} = \lambda_{0} L_{0} + \lambda_{1} L_{1} + \lambda_{2} L_{2} + \lambda_{3} L_{3}
\label{equ:5}
\end{equation}
where $\theta$ represents the parameters of the backdoored encoder, and $\lambda_{0}, \lambda_{1}, \lambda_{2}, \lambda_{3}$ are hyperparameters controlling the relative importance of each objective. We optimize this total loss using gradient descent to obtain the optimal parameters $\theta'$ for the backdoored encoder.The detailed training procedure is provided in Algorithm \ref{alg:fida}.

\subsection{Mechanistic Interpretation of Feature Instability}
To explain how  the proposed Instability Loss ($L_3$)  affects  perturbation-based detection, we provide a  local geometric interpretation  of the feature  mapping. This interpretation describes the intended behavior of the objective rather than providing a formal global robustness guarantee.

Traditional backdoor defenses  such as STRIP~\cite{gao2019strip} and SymND~\cite{zhu2025symnd}  exploit the tendency of conventional backdoor mappings to remain stable under  localized input perturbations.  This behavior can be represented through a small effective local sensitivity  in the vicinity of  a  triggered input:
\begin{equation}
\| f'(T(x) + \delta) - f'(T(x)) \| \leq K \| \delta \|
\end{equation}
where a relatively small  $K$  represents a locally stable feature response along the considered perturbation directions.

FIDA is  designed to weaken this stability  assumption through adversarial optimization. By minimizing  $L_3$  in Eq. (4), the attacker  reduces the cosine similarity between the unperturbed  triggered feature $f'(T(x))$ and its  noisy counterpart $f'(T(x)+\Delta)$. This encourages greater local sensitivity along the noise directions sampled during training. The resulting behavior is consistent with a larger effective  $K$  along those directions, although minimizing cosine similarity does not establish a global Lipschitz bound.

 Under noise-free conditions, $L_0$ maps the triggered sample toward  the  target representation. Under the finite perturbations used during training and evaluation, $L_3$ instead encourages the feature to move away from its unperturbed triggered representation. This conditional behavior reduces the perturbation-stability cue used by the evaluated  consistency-based  defenses. The loss ablation in the experimental section provides empirical evidence for this role of $L_3$.

\subsection{Implementation of Semantic Triggers}

\begin{figure}[!t]
    \centering
     \begin{minipage}{0.24\linewidth}
        \centering
         \includegraphics[width=\linewidth]{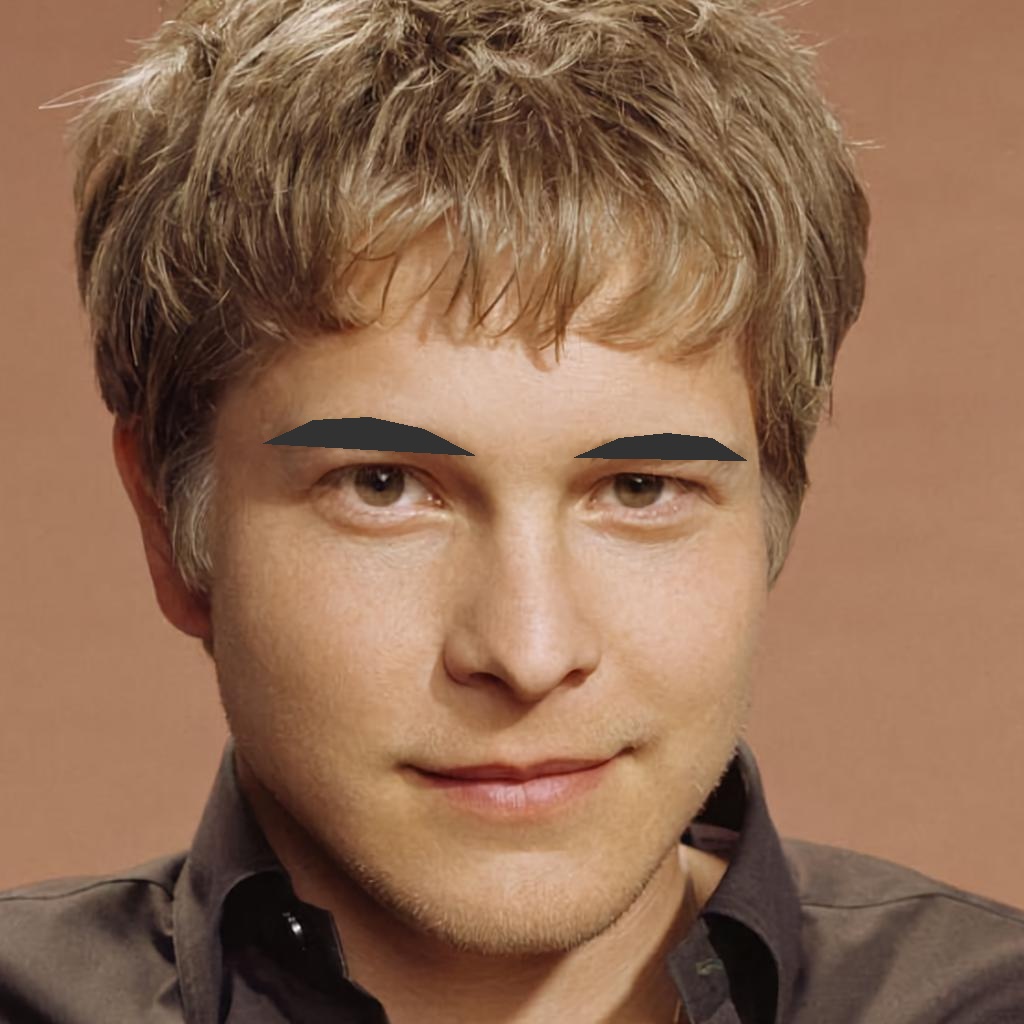}
        \centerline{(a) Eyebrow}
     \end{minipage}
    \hfill
    \begin{minipage}{0.24\linewidth}
        \centering
         \includegraphics[width=\linewidth]{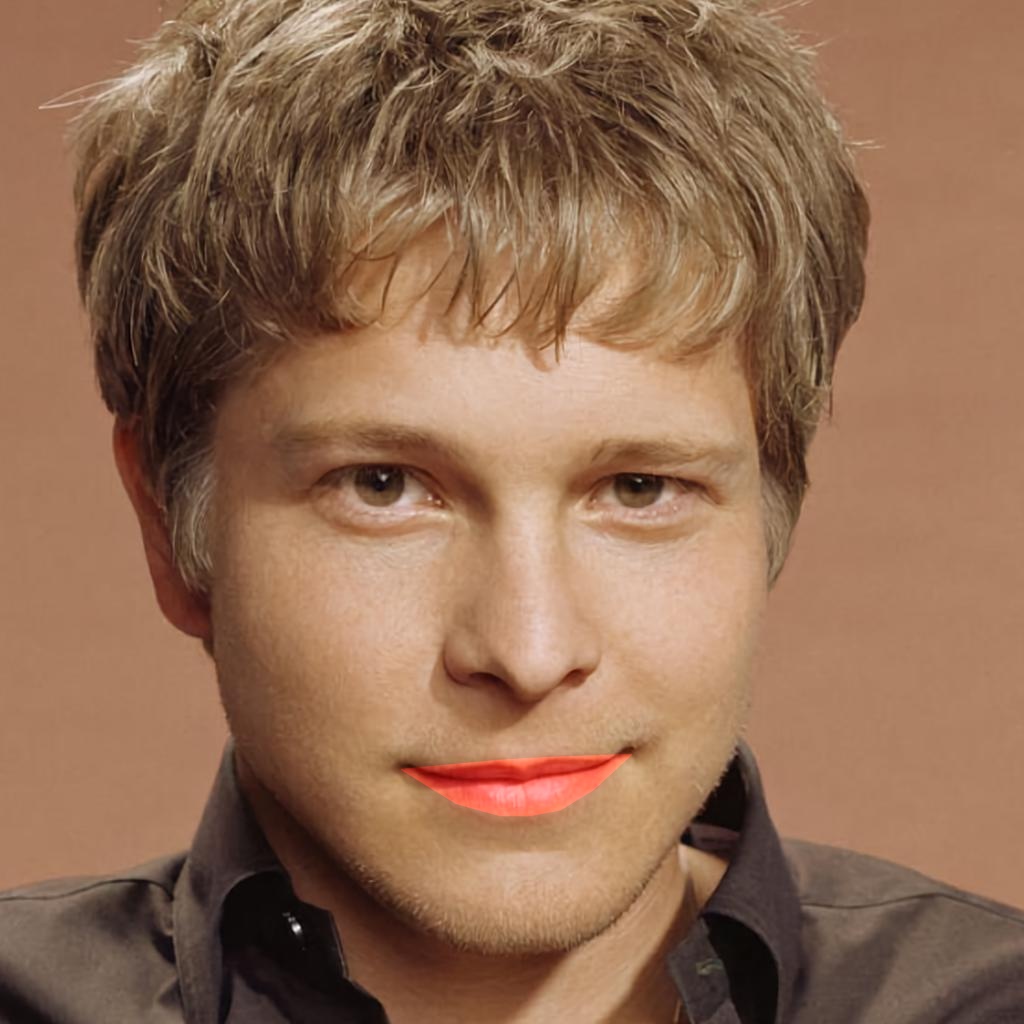}
        \centerline{(b) Red Lip}
     \end{minipage}
    \hfill
    \begin{minipage}{0.24\linewidth}
        \centering
         \includegraphics[width=\linewidth]{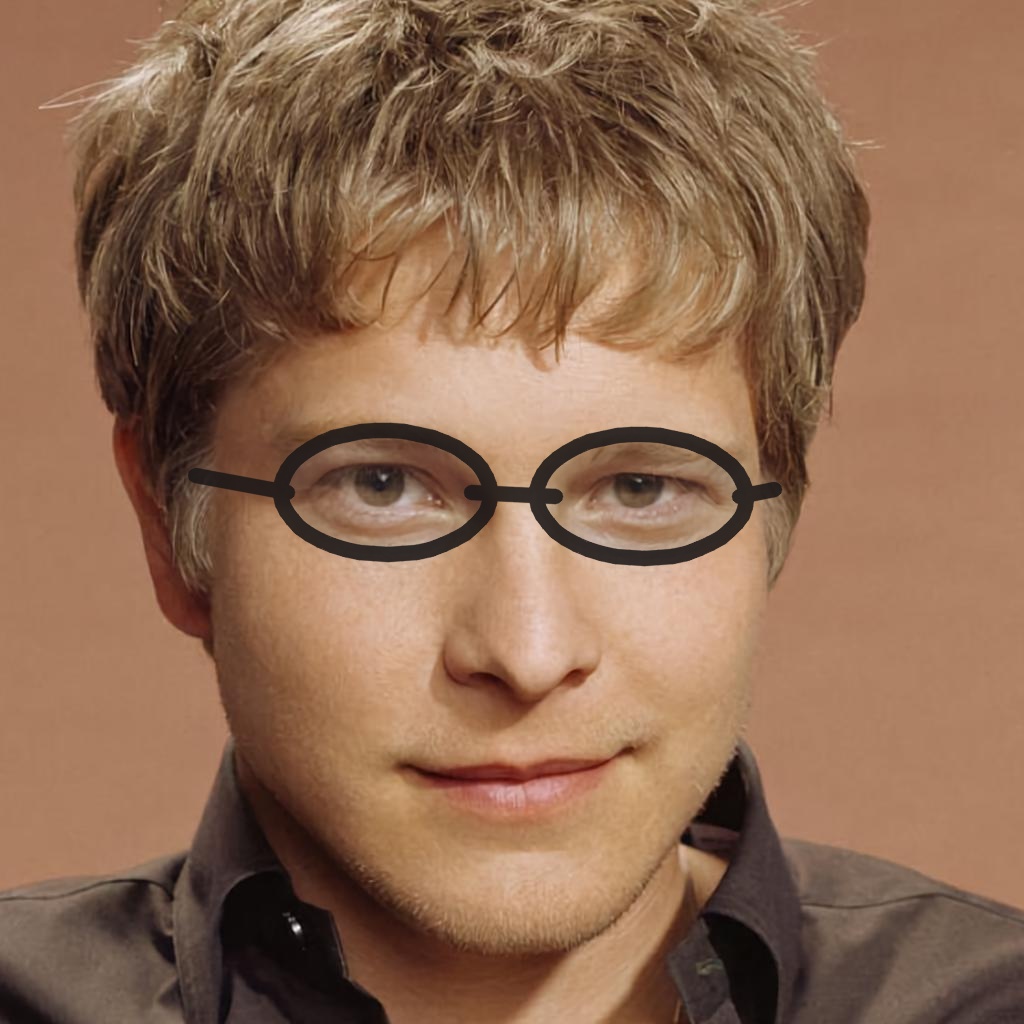}
        \centerline{(c) Glasses}
     \end{minipage}
    \hfill
    \begin{minipage}{0.24\linewidth}
        \centering
         \includegraphics[width=\linewidth]{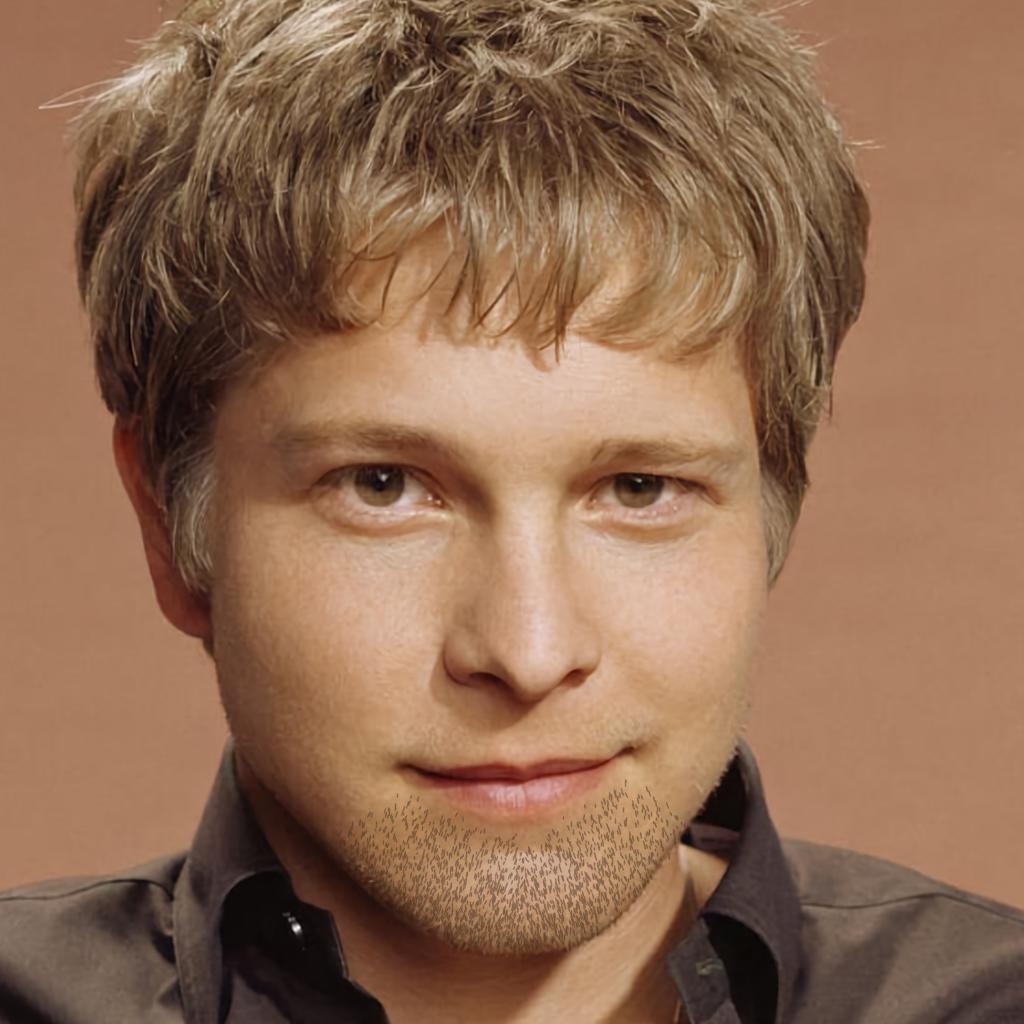}
        \centerline{(d) Beard}
     \end{minipage}
     \caption{Visualization of the four semantic triggers used in our evaluation: (a) eyebrow alteration, (b) red-lip coloring, (c) landmark-adaptive glasses, and (d) landmark-adaptive beard.}
     \label{fig:trigger}
\end{figure}

We implement four semantic triggers using Dlib's 68-point facial landmarks, as shown in Fig.~\ref{fig:trigger}. Their positions, scales, and shapes adapt to each input face. The \textbf{Eyebrow} trigger fills the convex hulls of landmarks 17--21 and 22--26 with dark gray to produce thickened eyebrows. The \textbf{Red Lip} trigger blends a red mask into the outer lip region defined by landmarks 48--59, with a transparency factor of 0.3. The \textbf{Glasses} trigger uses the eye landmarks 36--47 to determine the lens positions, inter-eye scale, and rotation. The \textbf{Beard} trigger constructs a textured lower-face region from the jaw, nose, and mouth landmarks, while adapting its color to the subject's facial appearance. Unlike fixed image-space patches, these triggers follow the facial geometry of each input. In particular, Glasses and Beard provide input-dependent accessory and facial-hair modifications with different locations and spatial coverage. For each trigger, we construct aligned clean, triggered, and triggered-plus-noise views. The noisy view is generated by adding Gaussian noise with a standard deviation of 30 to the triggered image, providing the paired inputs required by $L_3$.

\subsection{Complexity and Stability Analysis}
In this section, we discuss the computational overhead and optimization stability of FIDA.

\textbf{Computational Complexity.} Compared to the Naive Attack baseline, which also involves processing triggered counterparts for each benign sample, the additional computational burden of FIDA is primarily determined by the extra forward pass required to evaluate the Instability Loss $L_3$. Specifically, given a mini-batch of size $N$, FIDA requires an incremental forward propagation step for the noise-perturbed samples $T(x) + \Delta$. Since these operations---including Gaussian noise injection and similarity computation---are computationally inexpensive and fully parallelizable on modern GPU architectures, the training complexity remains within the same order of magnitude as the Naive Attack. Furthermore, the backdoored encoder $f'(\cdot)$ is initialized from a pre-trained model $f(\cdot)$, so malicious optimization begins from an already structured feature space rather than from random initialization.

\textbf{Optimization Stability.} A potential concern is whether the instability objective $L_3$, which encourages feature dispersion under perturbation, conflicts with the alignment objectives ($L_0, L_1, L_2$) and leads to gradient instability.

The Fidelity Losses ($L_1$ and $L_2$) constrain the attacked encoder to preserve the clean encoder's representations on sampled reference and shadow inputs, whereas $L_3$ acts on triggered inputs under sampled perturbations. Although these objectives do not formally guarantee gradient stability, the loss-removal and weight-sensitivity experiments provide empirical evidence that the combined objective preserves benign accuracy under the default setting.

\section{Evaluation}

\subsection{Experimental Setup}

\textbf{Datasets.} Our pipeline leverages four widely used facial analysis datasets spanning both pre-training and downstream evaluation stages:
\begin{itemize}
\item \textbf{WIKI\_CROP}~\cite{rothe2015dex}: A large-scale facial dataset with age annotations, collected from IMDb and Wikipedia. It contains over 60,000 images with metadata for age and gender, providing diverse real-world facial images for representation learning.
\item \textbf{FairFace}~\cite{karkkainen2021fairface}: A dataset designed to mitigate racial bias, containing 108,501 images balanced across 7 race groups (White, Black, Indian, East Asian, Southeast Asian, Middle Eastern, and Latino). We used FairFace to verify FIDA's performance on a distributionally balanced dataset.
\item \textbf{RAF-DB}~\cite{li2017reliable}: The Real-world Affective Faces Database contains great diversity in facial expressions. We used the basic emotion set (7 classes: Surprise, Fear, Disgust, Happiness, Sadness, Anger, Neutral) to evaluate the transferability of the backdoor to emotion recognition task.
\item \textbf{UTKFace}~\cite{zhang2017age}: A large-scale face dataset with wide variations in age (0-116 years), ethnicity, and pose. We utilized it for two specific classification tasks: race classification (5 classes: White, Black, Asian, Indian, Others) and age classification (discretized into 7 groups).
\end{itemize}

WIKI\_CROP and FairFace are used for self-supervised pre-training, while RAF-DB and UTKFace are reserved for downstream evaluation, covering tasks including emotion recognition, race classification, and age estimation. All images are aligned and resized to $64\times64$ pixels, and standard train-test splits are adopted when available; otherwise, datasets are randomly partitioned.

\textbf{Settings.} Following the experimental protocol from SymND~\cite{zhu2025symnd}, we first train a clean ResNet-18~\cite{he2016deep} encoder from scratch for 1000 epochs using SimCLR~\cite{chen2020simple} with random cropping, color jittering, and Gaussian blur. The attacker initializes the backdoored encoder from this clean encoder and maliciously fine-tunes it for 200 epochs on the complete WIKI\_CROP shadow split of 24,939 images using 10 fixed target-reference images. The clean encoder remains frozen during this stage. By default, we use the Eyebrow trigger and additive Gaussian noise with standard deviation 30. The victim subsequently freezes the supplied encoder and trains a three-layer MLP with two hidden layers of 512 and 256 units for 500 epochs on clean labeled data. The classifier uses Adam with a learning rate of 0.0001 and a batch size of 64.

Unless otherwise specified, the malicious fine-tuning stage optimizes only the four loss terms in Equation~\ref{equ:5}; it does not include an additional contrastive loss. We set $\lambda_0=\lambda_1=\lambda_2=\lambda_3=1$ and use SGD with a learning rate of 0.001, momentum of 0.9, weight decay of $5\times10^{-4}$, and batch size 32. Because FIDA modifies a supplied encoder rather than the victim's labeled dataset, it has no conventional downstream labeled-data poisoning ratio. Defense thresholds follow the original protocol of each defense and are calibrated only from its required clean calibration data, without using triggered test labels.

{\textbf{Baselines.} We compare FIDA with Naive Attack, BadEncoder~\cite{jia2022badencoder}, DRUPE~\cite{tao2024distribution}, and GhostEncoder~\cite{wang2024ghostencoder}. Naive Attack removes the feature-instability term $L_3$ from FIDA. BadEncoder learns a fixed-trigger-to-target representation mapping while preserving clean representations. DRUPE introduces distribution-preserving and assignment objectives to reduce the representation shift caused by poisoning. GhostEncoder uses a learned watermark generator to create input-dependent dynamic triggers. For the direct attack comparison, WIKI\_CROP is used upstream and UTKFace is used downstream. Every attacked encoder is trained for 200 epochs, and every downstream classifier is trained for 500 epochs. We retain the trigger and optimization protocol of each baseline.}

\textbf{Evaluation Metrics.} We evaluate utility, attack effectiveness, and defense performance. \textbf{Clean Accuracy (CA)} measures the clean encoder's downstream performance on benign images. \textbf{Benign Accuracy (BA)} measures the backdoored encoder's downstream performance on benign inputs. \textbf{Attack Success Rate (ASR)} is the percentage of triggered test inputs assigned to the attacker-selected target before defense. \textbf{Attack Success Rate After Defense (ASR-AD)} measures the same outcome after defensive filtering. True Positive Rate (TPR) is the proportion of attacks detected, while False Positive Rate (FPR) is the proportion of benign instances falsely flagged. True Negative Rate (TNR), equal to $1-\mathrm{FPR}$, is the proportion correctly accepted. Area Under the ROC Curve (AUROC) summarizes separation across thresholds.

\textbf{Experimental Environment.} The experiments were conducted on a server equipped with 4 NVIDIA RTX A5000 GPUs (24GB VRAM each). The implementation is based on PyTorch 2.4.1 with Python 3.8.20 and CUDA 12.1. Each training or evaluation job runs on a single GPU, while independent configurations are executed in parallel when multiple GPUs are available. Standard libraries including NumPy, OpenCV, and Kornia are used for data processing and augmentation.

\subsection{Main Results}

In this section, we evaluate the overall performance of the complete FIDA framework across diverse experimental scenarios. Table \ref{tab:main_results} presents a comprehensive summary of the results on two pre-training datasets transferred to three downstream classification tasks.

\begin{table}[t]
\centering
    \caption{Evaluation of attack effectiveness, utility, and defense evasion against SymND across diverse datasets.} 
    \label{tab:main_results}
    \resizebox{\textwidth}{!}{
      \begin{tabular}{cccccccc}
        \toprule
        \textbf{Pre-training} & \textbf{Downstream} & \textbf{Classification} & \multirow{2}{*}{\textbf{CA(\%)}} & \multirow{2}{*}{\textbf{Method}} & \multirow{2}{*}{\textbf{BA(\%)}} & \multirow{2}{*}{\textbf{ASR(\%)}} & \multirow{2}{*}{\textbf{ASR-AD(\%)}} \\
        \textbf{Dataset} & \textbf{Dataset} & \textbf{Task} & & & & &  \\
        \midrule
        \multirow{6}{*}{WIKI\_CROP} & \multirow{2}{*}{UTKFace} & \multirow{2}{*}{Race} & \multirow{2}{*}{76.29} & Naive Attack & 77.09 & 99.65 & 1.20 \\
        & & & & \textbf{FIDA} & 76.92 & 99.82 & 99.82 \\
        \cmidrule{2-8}
        & \multirow{2}{*}{RAF\_DB} & \multirow{2}{*}{Emotion} & \multirow{2}{*}{66.23} & Naive Attack & 66.40 & 97.84 & 1.30 \\
        & & & & \textbf{FIDA} & 66.72 & 99.10 & 99.10 \\
        \cmidrule{2-8}
        & \multirow{2}{*}{UTKFace} & \multirow{2}{*}{Age} & \multirow{2}{*}{63.56} & Naive Attack & 62.25 & 99.82 & 2.80 \\
        & & & & \textbf{FIDA} & 62.33 & 99.87 & 99.87 \\
        \midrule
        \multirow{6}{*}{FairFace} & \multirow{2}{*}{UTKFace} & \multirow{2}{*}{Race} & \multirow{2}{*}{79.73} & Naive Attack & 76.63 & 99.58 & 9.70 \\
        & & & & \textbf{FIDA} & 76.46 & 99.41 & 99.41 \\
        \cmidrule{2-8}
        & \multirow{2}{*}{RAF\_DB} & \multirow{2}{*}{Emotion} & \multirow{2}{*}{71.47} & Naive Attack & 66.20 & 0 & 0 \\
        & & & & \textbf{FIDA} & 65.48 & 97.47 & 97.47 \\
        \cmidrule{2-8}
        & \multirow{2}{*}{UTKFace} & \multirow{2}{*}{Age} & \multirow{2}{*}{65.73} & Naive Attack & 65.33 & 98.92 & 3.40 \\
        & & & & \textbf{FIDA} & 65.12 & 99.29 & 99.29 \\
        \bottomrule
      \end{tabular}
      }
\end{table}

\subsubsection{Effectiveness and Utility} First, we assess the primary goals of any backdoor attack: high success rate and minimal impact on benign functions. As shown in Table \ref{tab:main_results}, FIDA consistently achieves a near-perfect ASR, exceeding 99\% across almost all tested scenarios. Notably, FIDA maintains a high BA comparable to the CA of clean models. The degradation in benign performance is negligible, demonstrating that FIDA successfully implants a backdoor without compromising the encoder's general utility.

\begin{table}[t]
\centering
    
\caption{Comparison with representative backdoor attacks against self-supervised encoders using WIKI\_CROP upstream and UTKFace downstream.}
\label{tab:attack_baselines}
\begin{tabular}{lcc}
\toprule
\textbf{Method} & \textbf{BA (\%)} & \textbf{ASR (\%)} \\
\midrule
BadEncoder & 74.2670 & 92.5754 \\
DRUPE & 65.6402 & 97.0892 \\
GhostEncoder & 76.5239 & 61.8435 \\
\textbf{FIDA} & \textbf{76.9200} & \textbf{99.8200} \\
\bottomrule
\end{tabular}
\end{table}

Table~\ref{tab:attack_baselines} shows that FIDA achieves the highest ASR while retaining competitive BA. DRUPE approaches FIDA in ASR but substantially reduces BA, whereas GhostEncoder preserves BA but obtains a lower ASR. FIDA therefore provides the best BA--ASR balance in this setting.

\begin{figure*}[t]
    \centering

    \begin{minipage}{0.23\textwidth}
        \centering
        \includegraphics[width=1.0\linewidth]{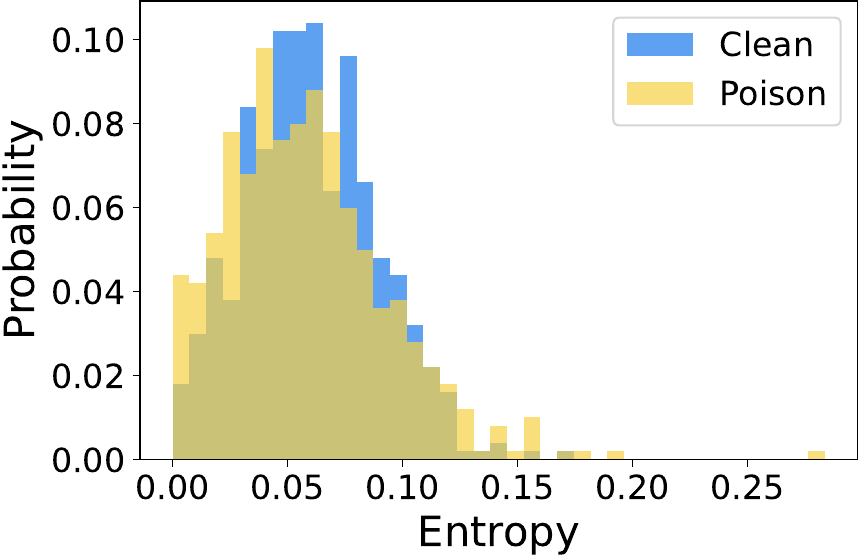}
        \centerline{(a) STRIP}
    \end{minipage}
    \hfill 
    \begin{minipage}{0.23\textwidth}
        \centering
        \includegraphics[width=1.0\linewidth]{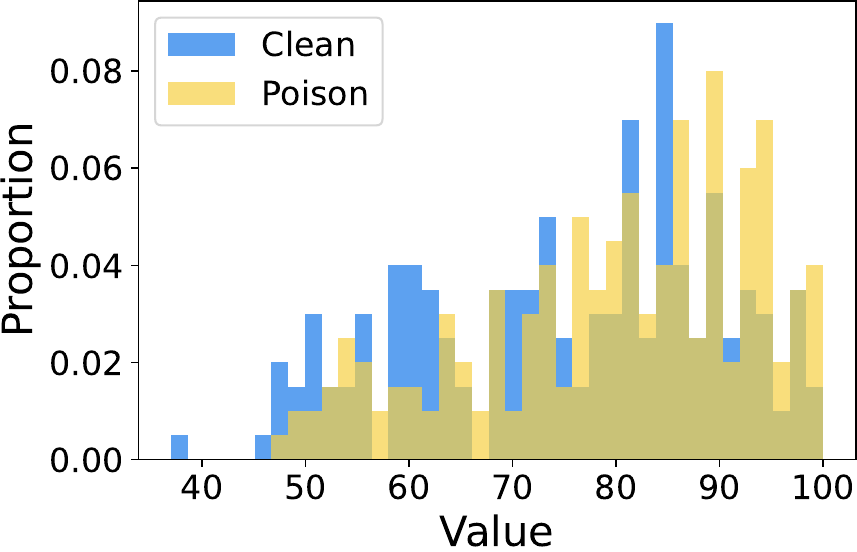}
        \centerline{(b) STRIP-CL}
    \end{minipage}
    \hfill 
    \begin{minipage}{0.23\textwidth}
        \centering
        \includegraphics[width=1.0\linewidth]{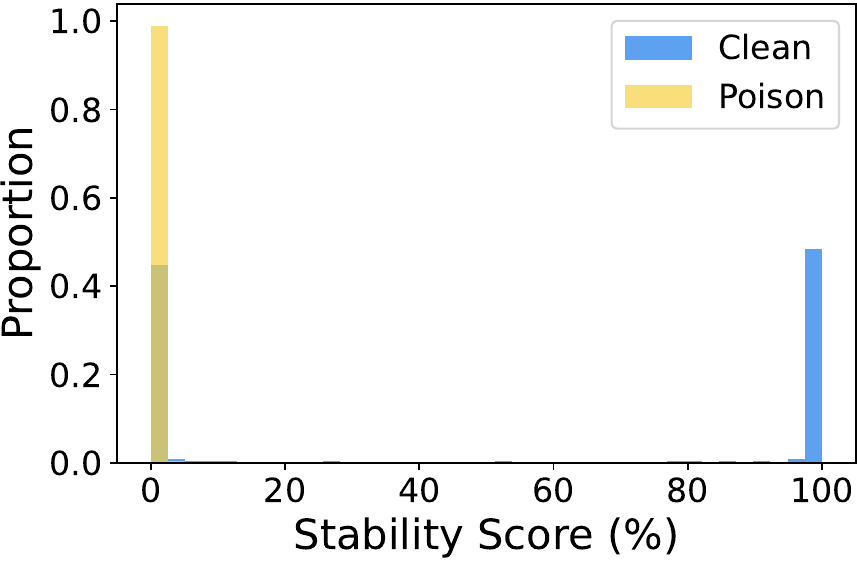}
        \centerline{(c) SymND}
    \end{minipage}
    \hfill 
    \begin{minipage}{0.23\textwidth}
        \centering
        \includegraphics[width=1.0\linewidth]{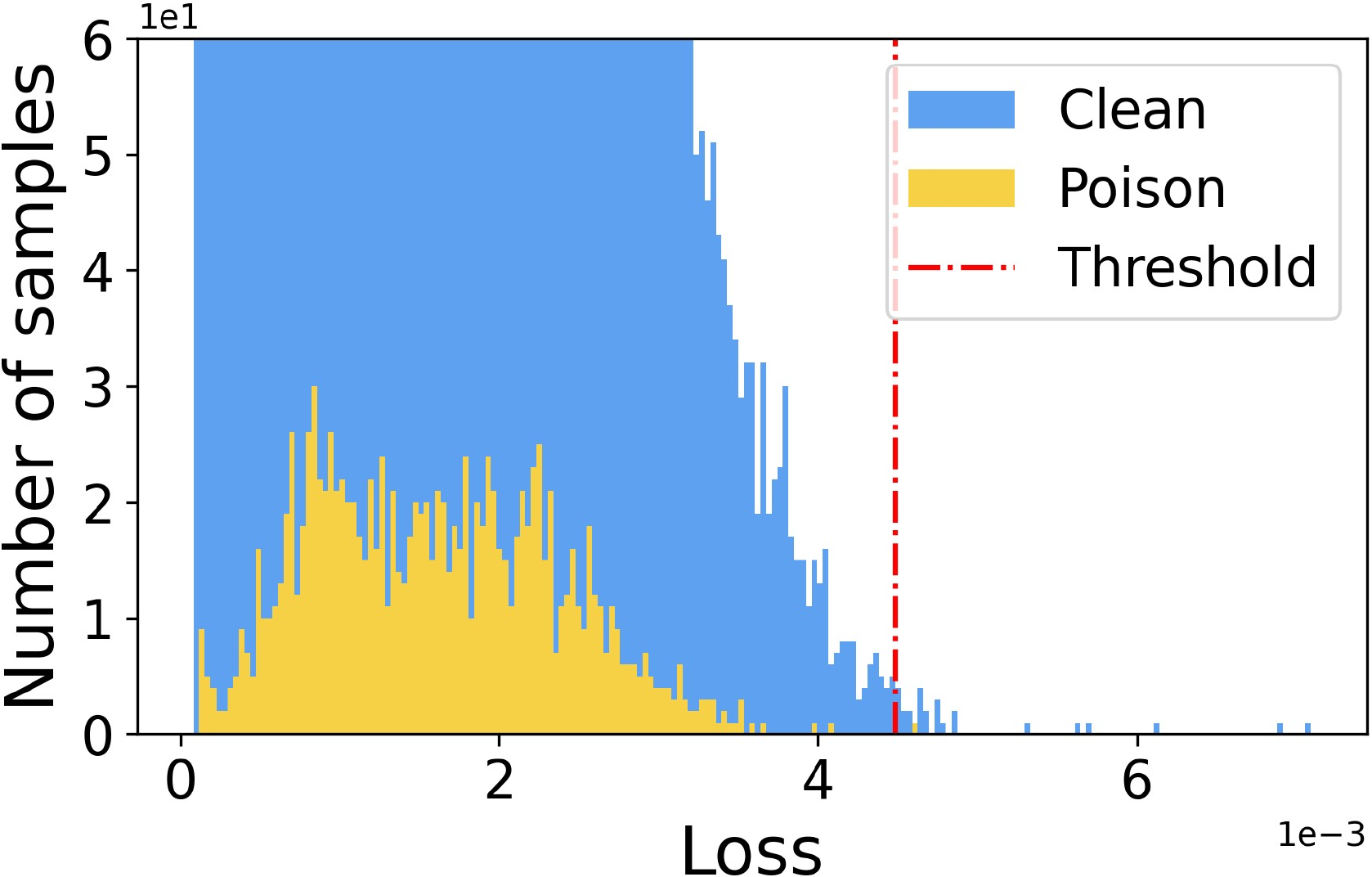}
        \centerline{(d) ASSET}
    \end{minipage}

    \caption{Evaluation of defense evasion against STRIP, STRIP-CL, SymND and ASSET}
    \label{fig:Defense Results}
\end{figure*}

\begin{figure}
    \centering
     \begin{minipage}{0.3\linewidth} 
            \centering
            \includegraphics[width=0.85\linewidth]{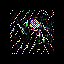}
            \vspace{2pt} 
            \scriptsize
            Clean Encoder\\
            $\mathcal{PL}^1$-Norm=0.1076\\  
            $\mathcal{L}^1$-Norm=1322.18    
        \end{minipage}
        \begin{minipage}{0.3\linewidth}
            \centering
            \includegraphics[width=0.85\linewidth]{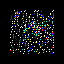}
            \vspace{2pt}
            \scriptsize
            Backdoored Encoder\\
            $\mathcal{PL}^1$-Norm=0.1149\\  
            $\mathcal{L}^1$-Norm=1411.98    
        \end{minipage}
    \caption{Evaluation of defense evasion against DECREE}
    \label{fig:DECREE}
\end{figure}

\subsubsection{Defense Evasion} We evaluate FIDA against representative perturbation-based, data-separation, and model-inversion defenses. The results demonstrate FIDA's strong ability to evade these different defense mechanisms.

\textbf{STRIP, STRIP-CL and SymND.} We first consider widely-used techniques designed to filter inputs at run-time. STRIP~\cite{gao2019strip} and its variant STRIP-CL~\cite{wang2024ghostencoder} operate by superimposing inputs with benign samples and measuring the entropy of prediction shifts, assuming backdoored inputs behave differently. We also evaluate against SymND~\cite{zhu2025symnd}, notably the first defense specifically designed to detect backdoors in self-supervised facial representations by analyzing feature stability under perturbation. The failure of entropy-based defenses is evident in Fig. \ref{fig:Defense Results} (a) (b) (c), where the entropy distributions of clean and backdoored samples completely overlap. Crucially, as reported in Table \ref{tab:main_results}, while Naive Attack is effectively neutralized by SymND (with ASR-AD dropping to a negligible level), FIDA's ASR-AD remains virtually identical to its original high ASR. This indicates that the induced feature instability successfully prevents these mechanisms from distinguishing and filtering out triggered samples.

\textbf{ASSET.} We further challenged FIDA against ASSET~\cite{pan2023asset}, a state-of-the-art defense that actively optimizes the encoder to maximize the loss variance of poisoned samples for separation. FIDA demonstrates robust evasion against this active separation mechanism. Under ASSET's standard adaptive thresholding, the detection yields a negligible True Positive Rate (TPR) of 0.08\% at a False Positive Rate (FPR) of 0.20\%, meaning over 99.9\% of the backdoor samples successfully evade sanitization. As shown in Fig. \ref{fig:Defense Results} (d), the normalized loss scores of FIDA-triggered and benign samples overlap substantially. This overlap prevents the Gaussian Mixture Model from establishing a valid separation boundary, rendering the defense ineffective in filtering out malicious inputs.

\begin{figure*}[t]
    \centering
    \begin{subfigure}[t]{0.35\textwidth}
        \centering
        \includegraphics[width=\linewidth]{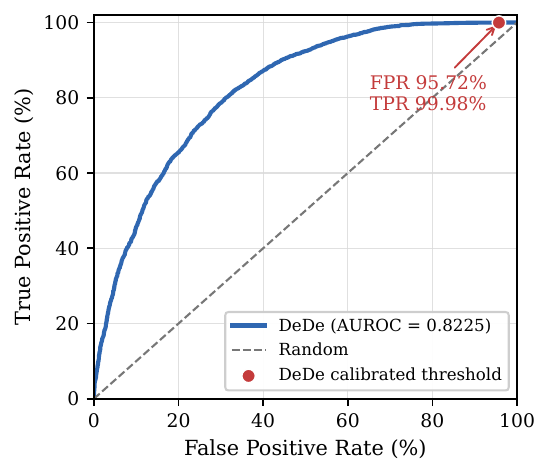}
        \caption{Complete ROC curve and calibrated operating point.}
        \label{fig:dede_full_roc}
    \end{subfigure}
    \begin{subfigure}[t]{0.35\textwidth}
        \centering
        \includegraphics[width=\linewidth]{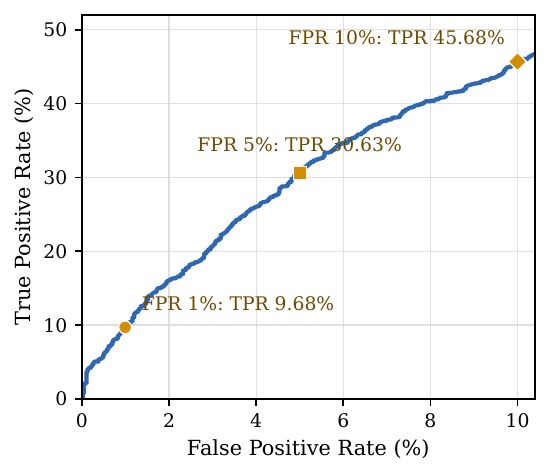}
        \caption{Detection performance in the low-FPR region.}
        \label{fig:dede_low_fpr}
    \end{subfigure}
    \caption{Detection performance of DeDe against FIDA. The calibrated threshold achieves a high TPR but produces an FPR of 95.72\%. Under practical FPR constraints, DeDe detects only a limited proportion of FIDA-triggered samples. Both curves are computed from the per-sample reconstruction scores.}
    \label{fig:dede_roc}
\end{figure*}

\textbf{DECREE.} We also evaluated FIDA against DECREE~\cite{feng2023detecting} (Fig. \ref{fig:DECREE}), a prominent static defense tailored for self-supervised encoders. DECREE inverts potential triggers and flags a model as backdoored if the $\mathcal{PL}^1$-Norm of the inverted trigger falls below a detection threshold $\tau=0.1$. Evaluation on the FIDA-poisoned encoder yields a $\mathcal{PL}^1$-Norm of 0.11. With this value surpassing $\tau$, DECREE erroneously classifies the FIDA encoder as ``benign''. This evasion suggests that FIDA's semantic triggers, which resemble natural facial features, violate the low-norm assumption of DECREE, thereby preventing the defense from isolating the backdoor pattern via inversion.

{\textbf{DeDe.} We adapt the recent decoder-based DeDe defense~\cite{hou2025dede} by training its decoder for 200 epochs on clean unlabeled WIKI\_CROP images. Following its reconstruction protocol, we use three independently sampled masks and set the detection threshold to 1.5 times the clean calibration mean. This produces a threshold of 5259.8383 without using triggered test labels. DeDe detects 4557 of 4558 triggered inputs, but it also rejects 4538 of 4741 clean inputs. As shown in Fig.~\ref{fig:dede_roc}(a), its 99.98\% TPR is accompanied by a 95.72\% FPR and a TNR of only 4.28\%. Although the overall AUROC is 0.8225, the calibrated operating point classifies almost all normal samples as suspicious. Fig.~\ref{fig:dede_roc}(b) further shows that, under FPR budgets of 1\%, 5\%, and 10\%, the corresponding TPR values decrease to 9.68\%, 30.63\%, and 45.68\%, respectively. Therefore, reducing the false-positive rate to a practically usable range causes most FIDA-triggered samples to evade detection. These results indicate that DeDe cannot effectively distinguish normal and FIDA-triggered samples in this experiment.}

{\textbf{BARBIE.} BARBIE~\cite{zhang2025barbie} is a recent model-level detector. It flags an encoder when its inverted Relative Competition Score indicators fall outside a normal range calibrated from known clean models. We follow BARBIE's original self-supervised protocol: a SimCLR ResNet-18 encoder, a two-layer classifier, OLR/ILR inversion, mean-centered RCS indicators, $\alpha{=}0.001$, $\gamma{=}0$, and $\omega_{\max}{=}0.5$. We retain its data-free, weights-only profiling and measure profile-level FPR by leave-one-encoder-out evaluation on unseen clean profiles. As shown in Table~\ref{tab:barbie}, every FIDA scenario raises an alarm. However, the same boundary also flags more than half of the clean profiles. BARBIE's min--max calibration assumes clean models from the same domain and cannot absorb profile differences across heterogeneous upstream domains. It therefore cannot reliably separate FIDA from normal encoders trained on different domains in the supply-chain setting.}

\begin{table}[t]
\centering
    
\caption{BARBIE under heterogeneously-sourced calibration ($\omega_{\max}{=}0.5$). FPR is measured via leave-one-encoder-out on unseen clean encoder profiles.}
\label{tab:barbie}
\begin{tabular}{lc}
\toprule
Downstream task & FPR \\
\midrule
UTKFace (Race)      & 53.3\% \\
RAF-DB (Emotion)    & 53.3\% \\
UTKFace (Age)       & 56.7\% \\
\bottomrule
\end{tabular}
\end{table}

\begin{table}[t]
\centering

\caption{Performance after applying MIMIC to FIDA encoders. BA-D and ASR-D denote benign accuracy and attack success rate after purification.}
\label{tab:mimic_defense}
\begin{tabular}{llccc}
\toprule
\textbf{Pre-training} & \textbf{Downstream} & \multirow{2}{*}{\textbf{CA (\%)}} & \multirow{2}{*}{\textbf{BA-D (\%)}} & \multirow{2}{*}{\textbf{ASR-D (\%)}} \\
\textbf{Dataset} & \textbf{Dataset} & & & \\
\midrule
\multirow{2}{*}{WIKI\_CROP} & UTKFace & 76.29 & 63.81 & 46.33 \\
\cmidrule(l){2-5}
& RAF-DB  & 66.23 & 55.11 & 2.71 \\
\midrule
\multirow{2}{*}{FairFace} & UTKFace & 79.73 & 65.91 & 44.54 \\
\cmidrule(l){2-5}
& RAF-DB  & 71.47 & 59.65 & 3.67 \\
\bottomrule
\end{tabular}
\end{table}

\textbf{MIMIC~\cite{han2025mutual}.} MIMIC treats the suspicious encoder as a teacher and trains a new student encoder on a small clean proxy set. A new downstream classifier must then be trained on the student representation. Following its protocol, we provide clean data equal to 4\% of the corresponding pre-training dataset. This procedure requires trusted upstream-domain data, teacher--student optimization, and downstream retraining, which adds substantial computation and limits its practicality for resource-constrained users. We evaluate four FIDA encoders trained with WIKI\_CROP or FairFace on UTKFace and RAF-DB. Table~\ref{tab:mimic_defense} shows that BA-D is 11.12--13.82 percentage points below CA in every setting, indicating substantial damage to benign utility. Although MIMIC reduces ASR-D, considerable residual attack success remains on UTKFace, while the stronger reduction on RAF-DB is accompanied by a benign-accuracy loss above 11 percentage points. MIMIC therefore does not achieve uniform, utility-preserving purification against FIDA.

\subsection{Ablation Study}

\subsubsection{Loss Components and Weights}

\begin{table}[t]
    \centering
    \caption{Ablation study on different loss terms using WIKI\_CROP upstream and UTKFace downstream.}
    \label{tab:ablation_study}
    \begin{tabular}{ccccc}
        \toprule
        \textbf{Removed Loss Terms} & \textbf{CA(\%)} & \textbf{BA(\%)} & \textbf{ASR(\%)} & \textbf{ASR-AD(\%)} \\
        \midrule
        $L_{0}$ & \multirow{5}{*}{76.29} & 76.71 & 45.30 & 45.30 \\
        $L_{1}$ &                        & 75.24 & 82.10 & 82.10 \\
        $L_{2}$ &                        & 59.61 & 57.42 & 57.42 \\
        $L_{3}$ &                        & 77.09 & 99.65 & 1.20  \\
        None &                        & 76.92 & 99.82 & 99.82 \\
        \bottomrule
    \end{tabular}
\end{table}

\begin{figure*}[t]
\centering
    
\includegraphics[width=\textwidth]{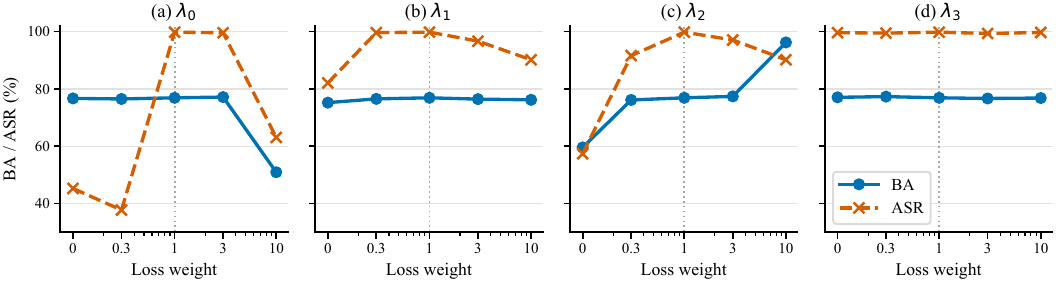}
\caption{Sensitivity to the four loss weights. Each panel varies one coefficient while fixing the other three to 1. Solid blue lines with circles denote BA, dashed orange lines with crosses denote ASR, and the vertical dotted line marks the default unit weight.}
\label{fig:lambda_sensitivity}
\end{figure*}

To isolate the contribution of each objective, we remove one loss term at a time and report the resulting utility, attack effectiveness, and post-defense effectiveness in Table~\ref{tab:ablation_study}. Removing $L_0$ weakens the trigger-to-target mapping, whereas removing $L_1$ produces a smaller loss of attack effectiveness. Removing $L_2$ damages both benign utility and attack effectiveness. This indicates that preserving clean representations also provides a stable feature space in which the target mapping can be learned.

The role of $L_3$ is different from those of the alignment and fidelity terms. Removing it leaves benign utility and the undefended attack largely intact, but makes the attack ineffective after SymND. Thus, $L_3$ primarily supports defense evasion by weakening the feature-stability cue rather than establishing the basic backdoor mapping.

We further vary all four loss coefficients while fixing the other three, as shown in Fig.~\ref{fig:lambda_sensitivity}. The extended sweep shows that attack alignment is sensitive to insufficient or excessive weighting, while an overly large clean-feature weight makes benign preservation dominate the attack objective. The instability weight is comparatively tolerant once activated, consistent with its specialized role in defense evasion. We therefore retain unit weights as the balanced default.

\subsubsection{Noise Intensity}

\begin{table}[t]   
    \centering
    \caption{Impact of noise intensity using WIKI\_CROP upstream and UTKFace downstream.}
    \label{tab:noise_sensitivity}
    \begin{tabular}{cccccc}
        \toprule
        \textbf{\multirow{2}{*}{Noise Std Dev}} & \textbf{\multirow{2}{*}{BA(\%)}} & \textbf{\multirow{2}{*}{ASR(\%)}} & \multicolumn{3}{c}{\textbf{ASR-AD(\%)}} \\
        \cmidrule(lr){4-6}
        & & & \textbf{STRIP} & \textbf{STRIP-CL} & \textbf{SymND} \\
        \midrule
        10 & 76.17 & 99.50 & 99.50 & 99.50 & 99.50 \\
        20 & 76.08 & 98.03 & 98.03 & 98.03 & 98.03 \\
        30 & 76.92 & 99.82 & 99.82 & 99.82 & 99.82 \\
        40 & 76.55 & 99.71 & 99.71 & 99.71 & 99.71 \\
        50 & 77.29 & 99.65 & 99.65 & 99.65 & 99.65 \\
        \bottomrule
    \end{tabular}
\end{table}

To examine whether FIDA depends on a narrowly selected perturbation magnitude, we vary the Gaussian-noise standard deviation used during instability training; the results are summarized in Table~\ref{tab:noise_sensitivity}. Benign utility and attack effectiveness remain stable across the tested range, and the post-defense results closely track the undefended attack. This consistency suggests that the instability objective does not require a precisely tuned noise intensity. We therefore retain the intermediate setting $\sigma=30$ as the default.

\subsubsection{Physical Degradations}
\label{subsec:physical_degradation}

{Following the robustness-test design of INACTIVE~\cite{zhang2025invisible}, we independently apply Gaussian blur, JPEG compression, Poisson noise, and salt-and-pepper noise to the clean and triggered UTKFace test images. The attacked encoder remains fixed, and the downstream classifier follows the same 500-epoch training protocol as the standard evaluation. ASR excludes samples whose ground-truth label is already the target class, preventing ordinary correct target predictions from being counted as successful attacks. The undegraded reference obtains 76.92\% BA and 99.62\% strict ASR.}

\begin{figure*}[t]
\centering
    
\includegraphics[width=\textwidth]{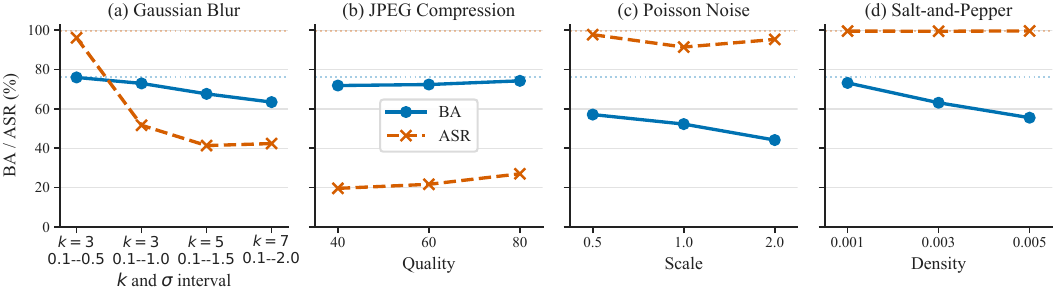}
\caption{Robustness of FIDA under simulated physical-world image degradations using WIKI\_CROP upstream and UTKFace downstream. Each panel varies the strength of one degradation. For Gaussian blur, each tick reports the kernel size $k$ and the sampled $\sigma$ interval. Solid blue lines with circles denote BA, dashed orange lines with crosses denote ASR, and faint dotted horizontal lines denote the corresponding undegraded references.}
\label{fig:physical_degradation}
\end{figure*}

To determine how deployment-like corruption affects FIDA, we compare the degradation trends in Fig.~\ref{fig:physical_degradation}. Mild blur largely preserves both utility and attack effectiveness, whereas stronger smoothing weakens the localized semantic evidence used by the trigger before it removes the broader facial information needed for benign classification. JPEG recompression produces a different imbalance: benign features remain comparatively usable, but fine trigger-related details are strongly suppressed. Salt-and-pepper and Poisson noise mainly reduce benign utility while leaving the target mapping more stable. FIDA is therefore more resilient to stochastic pixel corruption than to transformations that smooth or compress semantic facial details.

\subsubsection{Generalization}

\begin{table}[t]
    \centering
    \caption{Generalization across SSL frameworks using WIKI\_CROP upstream and UTKFace downstream.}
    \label{tab:arch_generalization}
    \begin{tabular}{ccccccc}
        \toprule
        \textbf{\multirow{2}{*}{Architecture}} & \textbf{\multirow{2}{*}{CA(\%)}} & \textbf{\multirow{2}{*}{BA(\%)}} & \textbf{\multirow{2}{*}{ASR(\%)}} & \multicolumn{3}{c}{\textbf{ASR-AD(\%)}} \\
        \cmidrule(lr){5-7}
        & & & & \textbf{STRIP} & \textbf{STRIP-CL} & \textbf{SymND} \\
        \midrule
        SimCLR & 76.29 & 76.92 & 99.82 & 99.82 & 99.82 & 99.82 \\
        BYOL & 75.13 & 75.79 & 87.74 & 87.74 & 87.74 & 87.74 \\
        MoCo v2 & 76.46 & 75.01 & 80.04 & 80.04 & 80.04 & 80.04 \\
        \bottomrule
    \end{tabular}
\end{table}

To examine whether FIDA depends on a particular self-supervised objective, we implement it with SimCLR, BYOL, and MoCo v2 under the same upstream and downstream setting; the results are reported in Table~\ref{tab:arch_generalization}. These frameworks represent contrastive learning, asymmetric prediction, and momentum-based representation learning. Benign utility remains comparable across them, while attack effectiveness is lower with BYOL and MoCo v2 than with SimCLR. Their different update rules and feature geometries may change how quickly a shared attack configuration establishes trigger-to-target alignment. Nevertheless, ASR-AD remains equal to the corresponding ASR for all three frameworks, indicating that the instability-based evasion mechanism is not tied to SimCLR.

We next test whether the attack transfers across different backbone architectures. Table~\ref{tab:backbone_generalization} compares three backbones: the original ResNet-18, the deeper ResNet-34, and VGG-16-BN, which has no residual connections. ResNet-34 preserves a stronger utility--effectiveness balance than VGG-16-BN. Residual connections may help retain pretrained representations during malicious fine-tuning, whereas VGG-16-BN learns a different feature hierarchy under the same loss weights. The unchanged relationship between ASR and ASR-AD nevertheless shows that the evasion mechanism transfers beyond the residual family. Backbone-specific tuning may further improve the balance on VGG-16-BN.

\begin{table}[t]
    \centering
    
    \caption{Evaluation with different encoder backbones using WIKI\_CROP upstream and UTKFace downstream.}
    \label{tab:backbone_generalization}
    \begin{tabular}{lccc}
\toprule
\textbf{Backbone} & \textbf{BA (\%)} & \textbf{ASR (\%)} & \textbf{ASR-AD (\%)} \\
\midrule
ResNet-18  & 76.92 & 99.82 & 99.82 \\
ResNet-34  & 73.95 & 92.61 & 92.61 \\
VGG-16-BN  & 67.05 & 84.71 & 84.71 \\
\bottomrule
    \end{tabular}
\end{table}

\begin{table}[t]
    \centering
    \caption{Generalization across semantic triggers using WIKI\_CROP upstream and UTKFace downstream.}
    \label{tab:trigger_generalization}
    \begin{tabular}{ccccccc}
        \toprule
        \textbf{\multirow{2}{*}{Trigger}} & \textbf{\multirow{2}{*}{CA(\%)}} & \textbf{\multirow{2}{*}{BA(\%)}} & \textbf{\multirow{2}{*}{ASR(\%)}} & \multicolumn{3}{c}{\textbf{ASR-AD(\%)}} \\
        \cmidrule(lr){5-7}
        & & & & \textbf{STRIP} & \textbf{STRIP-CL} & \textbf{SymND} \\
        \midrule
        Eyebrow & \multirow{4}{*}{76.29} & 76.92 & 99.82 & 99.82 & 99.82 & 99.82 \\
        Lip &   & 76.65 & 98.71 & 98.71 & 98.71 & 98.71 \\
        Glasses & & 75.49 & 96.07 & 96.07 & 96.07 & 96.07 \\
        Beard & & 75.76 & 81.86 & 81.86 & 81.86 & 81.86 \\
        \bottomrule
    \end{tabular}
\end{table}

Finally, we examine whether FIDA relies on one specific semantic appearance by evaluating four triggers with different locations, structures, and spatial coverage; the results are presented in Table~\ref{tab:trigger_generalization}. Eyebrow and Red Lip are localized cosmetic modifications applied to different facial regions, and their similar behavior shows that the target mapping is not tied to one position or color change. Glasses introduce an input-adaptive accessory whose geometry follows each face, yet the attack remains effective. Beard modifies a broader and more variable lower-face region, making a consistent target mapping harder to learn. These results show that FIDA supports cosmetic, accessory-based, and facial-hair triggers, while also revealing that trigger variability and coverage influence attack effectiveness.

\begin{figure*}[t]
\centering
    
\includegraphics[width=0.7\textwidth]{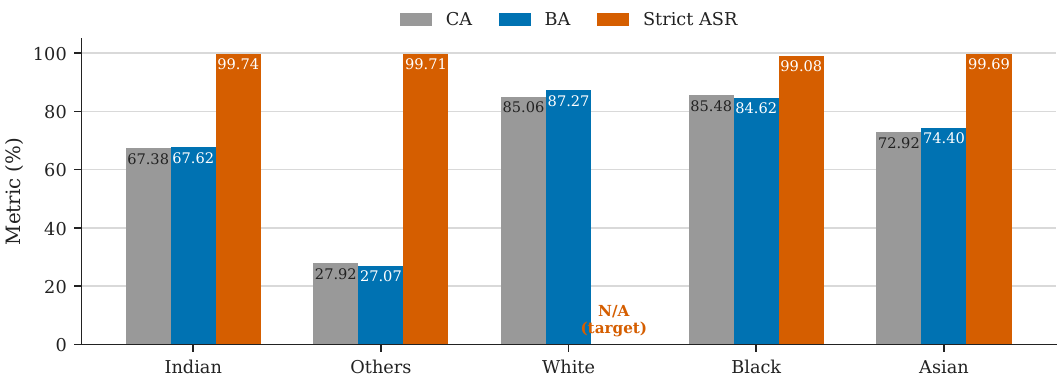}
\caption{Race-group analysis on UTKFace. Gray, blue, and orange bars denote CA, BA, and strict ASR, respectively. Strict ASR measures the proportion of triggered non-target samples classified as the target class. Since White is the attack target, its ASR is undefined and marked as N/A.}
\label{fig:race_group_fairness}
\end{figure*}

\begin{figure}[t] 
    \centering

    \includegraphics[width=0.55\linewidth]{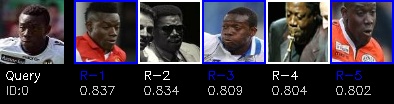}

    \includegraphics[width=0.55\linewidth]{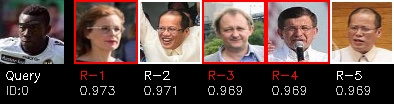}

    \caption{A qualitative face-retrieval example on VGGFace2. Top: a clean query. Bottom: the same query with a trigger.}
    \label{fig:face_retrieval}
    \vspace{-0.5cm}
\end{figure}

\subsection{Demographic Bias and Fairness Implications}
\label{subsec:demographic_subgroup_analysis}

Semantic facial modifications may interact with demographic differences in facial appearance. We therefore conduct a subgroup analysis using the five race labels in UTKFace. FIDA is trained for 200 epochs using WIKI\_CROP as the upstream dataset. The downstream classifier is trained for 500 epochs on UTKFace. The clean encoder provides CA. The FIDA encoder provides BA and ASR. We compute CA and BA on 4,741 clean test images. We compute strict ASR on 4,558 triggered images with their original race labels. White is the attack target. We therefore report ASR only for the four non-target groups.

As shown in Fig.~\ref{fig:race_group_fairness}, strict ASR ranges from 99.08\% to 99.74\% across the four non-target groups. The gap is only 0.66 percentage points. The group-wise difference between BA and CA ranges from $-0.87$ to $+2.21$ percentage points. Others has the lowest CA and BA. This UTKFace label is a heterogeneous catch-all category. It also has the smallest clean test subset, with 351 images. Its diverse composition and limited sample size make it harder to classify. Its BA is 27.07\%, close to its CA of 27.92\%. Thus, FIDA does not cause the low accuracy of this group.

The results do not show a pronounced race-group difference in ASR. They also do not show that FIDA amplifies the existing group differences in utility. This analysis uses one target class and the race labels available in UTKFace. It does not cover intersectional attributes or the natural occurrence of semantic facial features. We will examine these factors with additional target classes and demographic annotations in future work.

\subsection{Generalization to Metric-Based Retrieval}
\label{sec:face_retrieval}

{We conduct a cross-domain, zero-shot retrieval experiment on VGGFace2~\cite{cao2018vggface2} using its identity and demographic annotations~\cite{greco2020benchmarking}. VGGFace2 contains 3.31 million face images from 9,131 identities, with substantial variation in pose, age, illumination, and ethnicity. We use 1\% of its identities in this experiment. The WIKI\_CROP-pretrained clean and FIDA encoders are frozen to rank a disjoint query--gallery split by cosine similarity. Rank-$k$ measures whether the top-$k$ results contain the query identity, while ASR@$k$ measures whether they contain the attacker-selected target group (\emph{White}). Figure~\ref{fig:face_retrieval} and Table~\ref{tab:retrieval_results} report the qualitative and quantitative results.}

\begin{table}[t]
    \centering
    
    \caption{Zero-shot face retrieval results on VGGFace2. Both encoders are pre-trained on WIKI\_CROP and frozen during retrieval. Higher identification rates indicate better clean utility, while higher ASR indicates a more successful targeted attack.}
    \label{tab:retrieval_results}
    \small
    \setlength{\tabcolsep}{4pt}
    \begin{tabular}{llcccc}
        \toprule
        \multirow{2}{*}{Model} &
        \multirow{2}{*}{Query} &
        \multicolumn{2}{c}{Identification Rate (\%)} &
        \multicolumn{2}{c}{Target ASR (\%)} \\
        \cmidrule(lr){3-4} \cmidrule(lr){5-6}
        & & Rank-1 & Rank-5 & ASR@1 & ASR@5 \\
        \midrule
        Clean encoder & Clean     & 50.88 & 75.00 & -- & -- \\
        Clean encoder & Triggered & -- & -- & 11.67 & 48.67 \\
        FIDA encoder  & Clean     & 43.88 & 69.38 & -- & -- \\
        FIDA encoder  & Triggered & -- & -- & \textbf{55.83} & \textbf{93.50} \\
        \bottomrule
    \end{tabular}
\end{table}

{Relative to the clean-encoder control, FIDA increases ASR by 44.16 and 44.83 percentage points at Rank~1 and Rank~5, with ASR@5 reaching 93.50\%. Clean Rank-1 and Rank-5 identification decrease by 7.00 and 5.62 points, indicating that the targeted shift transfers to cross-domain retrieval with a moderate utility cost.}

\section{Discussion and Open Challenges}

The experimental results show that FIDA can manipulate self-supervised facial encoders while preserving their downstream utility in most evaluated settings. They also reveal several broader questions that cannot be fully addressed by classification accuracy and attack success rate alone. In this section, we discuss possible defenses against feature instability, the challenges of physical-world deployment, and the demographic and societal implications of semantic facial triggers.

\subsection{Potential Countermeasures Against Feature Instability}

FIDA is designed to manipulate the feature-stability assumption used by perturbation-based defenses. The defense results suggest that relying on a single stability cue may be insufficient when the encoder itself has been maliciously optimized against that cue. A stronger defense may need to combine several sources of evidence, such as representation consistency, reconstruction behavior, temporal information, and downstream prediction patterns.

Temporal consistency provides one possible direction. Facial analysis is often applied to video rather than isolated images. Although a semantic trigger may remain visually consistent across adjacent frames, a compromised encoder may produce abnormal changes in its representations when pose, illumination, or image quality varies. A defense could therefore examine feature trajectories across consecutive frames instead of making a decision from one perturbed image. Abrupt representation changes that are inconsistent with the observed facial motion may provide additional evidence of abnormal encoder behavior.

Multi-modal consistency provides another direction. Multimedia applications may jointly process facial appearance, speech, and contextual information. For example, if a visual trigger changes an emotion prediction to ``Angry,'' the system could compare this output with acoustic and temporal cues. A persistent disagreement between modalities would not by itself prove the presence of a backdoor, but it could initiate further inspection. These directions require video or multi-modal benchmarks and remain open problems beyond the current image-based evaluation.

\subsection{Complexities in Physical-World Multimedia Deployment}

As evaluated in Section~\ref{subsec:physical_degradation}, FIDA remains effective under mild Gaussian blur and the tested salt-and-pepper and Poisson noise conditions. However, Figure~\ref{fig:physical_degradation} also shows that stronger blur and JPEG compression substantially reduce ASR. These results indicate that the robustness of semantic triggers depends on how a degradation changes their facial appearance. Local pixel noise often preserves the main semantic structure, whereas blur and compression may weaken the visual details used to establish the trigger-to-target mapping.

These controlled degradations approximate several effects introduced during image capture, storage, and transmission, but they do not fully reproduce a physical environment. Real makeup and wearable accessories may appear differently across cameras, illumination conditions, facial poses, viewing distances, and video-compression pipelines. Evaluating these factors with physically reproduced triggers is therefore an important extension of the current simulated experiments.

\subsection{Demographic Bias, Fairness, and Societal Impact}
\label{subsec:demographic_fairness}
Section~\ref{subsec:demographic_subgroup_analysis} reports CA, BA, and ASR separately across the five race groups in UTKFace. As shown in Fig.~\ref{fig:race_group_fairness}, attack effectiveness and benign performance are not completely uniform across the evaluated groups. The lower accuracy of the ``Other'' group may be related to the heterogeneous racial identities represented by this broad category. The subgroup results therefore provide a more informative view than aggregate BA and ASR alone.

Semantic facial triggers create an additional fairness concern because eyebrow appearance, lip color, glasses, and facial hair can occur naturally. Their prevalence and appearance may vary with age, gender expression, cultural practices, skin tone, and facial structure. Some users may therefore naturally resemble a trigger without intentionally presenting one. The subgroup experiment measures differences in attack effectiveness, but it does not yet measure the frequency of such accidental activation.

The consequences may also depend on the downstream application. In demographic classification, a targeted backdoor could systematically map members of one group to another category, corrupting demographic statistics or decisions based on them. In identity verification, access control, telehealth, or automated assessment, unequal attack activation could result in denial of service, inaccurate evaluation, or disproportionate harm to particular populations.

Defenses should not treat glasses, facial hair, makeup, or eyebrow appearance themselves as evidence of malicious behavior. Otherwise, ordinary facial attributes could become a source of group-dependent false alarms. A safer direction is to detect abnormal model behavior rather than judging the semantic attribute. Future work will examine accidental activation rates, intersections among race, age, and gender, and group-specific false-positive rates of backdoor defenses.

\section{Conclusion}

Current backdoor attacks are often easily caught by advanced defenses. To solve this problem, we propose FIDA, a novel backdoor attack framework. By introducing a new feature instability mechanism, FIDA trains the encoder to  increase the sensitivity of triggered features along perturbation directions sampled during attack optimization . Our comprehensive evaluations across different datasets, model architectures, and trigger types show that FIDA achieves high attack success rates while maintaining the model's normal performance. It successfully evades  several evaluated  defense mechanisms.

For future work, we plan to explore several directions. First, we aim to extend the FIDA framework from the digital domain to physical world attacks. Since our semantic triggers correspond to natural facial features, we plan to implement them using physical mediums, such as makeup or accessories. Second, we will study how to design stronger defenses that can detect backdoor behaviors hidden by this induced instability.

\begin{acks}
We are deeply grateful to the anonymous reviewers for their thoughtful suggestions and rigorous reviews that helped refine our research.
\end{acks}

\bibliographystyle{ACM-Reference-Format}
\bibliography{sample-base}

\end{document}